\documentclass[conference]{IEEEtran}
\IEEEoverridecommandlockouts

\usepackage{cite}
\usepackage{amsmath,amssymb,amsfonts}
\usepackage{algorithmic}
\usepackage{graphicx}
\usepackage{textcomp}
\usepackage{xcolor}
\usepackage{comment}

\usepackage[table]{xcolor}

\usepackage{adjustbox}
\usepackage{mwe}

\usepackage{makecell}
\usepackage{diagbox}
\usepackage{tikz}
\usepackage{multirow}
\usepackage{hyperref}
\usepackage{booktabs}

\usepackage{siunitx}

\usepackage[a4paper,margin=2cm]{geometry}
\usepackage{graphicx}
\usepackage{caption}
\usepackage{subcaption}
\usepackage{array}
\usepackage{booktabs}
\usepackage{hyperref} 

\usepackage{grffile}
 
\graphicspath{{figures/}}
 
\newcommand{\gridimg}[1]{%
    \includegraphics[width=\linewidth,height=3.4cm]{#1}%
}

\usepackage{listings}
\usepackage{color}

\definecolor{dkgreen}{rgb}{0,0.6,0}
\definecolor{gray}{rgb}{0.5,0.5,0.5}
\definecolor{mauve}{rgb}{0.58,0,0.82}

\definecolor{green}{rgb}{0,0.5,0}
\definecolor{red}{rgb}{0.7,0,0}

\newcolumntype{G}{>{\centering\arraybackslash}m{0.24\linewidth}}

\hypersetup{
    colorlinks=true,       
    linkcolor=blue,        
    citecolor=black,         
    filecolor=magenta,     
    urlcolor=blue          
}

\usepackage{stfloats}

\newsavebox{\tempbox}

\usepackage{float}

\def\BibTeX{{\rm B\kern-.05em{\sc i\kern-.025em b}\kern-.08em
T\kern-.1667em\lower.7ex\hbox{E}\kern-.125emX}}

\begin{document}

\newcommand{\sceneA}{\detokenize{./resources/qualitative_results/00599_rgb_000599_id600_ts1774982988990138952/}}
\newcommand{\sceneB}{\detokenize{./resources/qualitative_results/04699_rgb_000050_id51_ts1774983234415047800/}}
\newcommand{\sceneC}{\detokenize{./resources/qualitative_results/05979_rgb_000655_id656_ts1774983318490465984/}}
\newcommand{\sceneD}{\detokenize{./resources/qualitative_results/11119_rgb_000026_ts1775933026984364176/}}
\newcommand{\sceneE}{\detokenize{./resources/qualitative_results/09959_rgb_002183_ts1775932958766946672/}}
\newcommand{\sceneF}{\detokenize{./resources/qualitative_results/09099_rgb_001323_ts1775932914844517112/}}


\title{Stereo LWIR-RGB Automatic Labeling for Low-Light Camera-Based VRU Detection

\thanks{We acknowledge the funding provided by Henry Ford Foundation Finland and the computational resources provided by the Aalto Science-IT project.}}

\author{
\IEEEauthorblockN{1\textsuperscript{st} Dimitrios Bouzoulas}
\IEEEauthorblockA{\textit{Energy and Mechanical Engineering} \\
\textit{Aalto University}\\
Espoo, Finland \\
dimitrios.bouzoulas@aalto.fi}
\and
\IEEEauthorblockN{2\textsuperscript{nd} Jäppinen Pyry}
\IEEEauthorblockA{\textit{Energy and Mechanical Engineering} \\
\textit{Aalto University}\\
Espoo, Finland \\
pyry.jappinen@aalto.fi}

\and
\IEEEauthorblockN{3\textsuperscript{rd} Pejman Habibiroudkenar}
\IEEEauthorblockA{\textit{Energy and Mechanical Engineering} \\
\textit{Aalto University}\\
Espoo, Finland \\
pejman.habibiroudkenar@aalto.fi}

\and
\IEEEauthorblockN{4\textsuperscript{rd} Eerik Alamikkotervo}
\IEEEauthorblockA{\textit{Energy and Mechanical Engineering} \\
\textit{Aalto University}\\
Espoo, Finland \\
eerik.alamikkotervo@aalto.fi}

\and
\IEEEauthorblockN{5\textsuperscript{th} Risto Ojala}
\IEEEauthorblockA{\textit{Energy and Mechanical Engineering} \\
\textit{Aalto University}\\
Espoo, Finland \\
risto.j.ojala@aalto.fi}
}

\maketitle

\begin{abstract}

VRU detection in low-light conditions is a key component in ensuring VRU safety during the most collision prone time of the day. Currently, advances in low-light detection are limited due to the lack of large scale datasets and automated labeling methods. Most autolabeling methods use RGB detections as their labels source, which cannot be replicated in low-light conditions. To fill this gap in this paper we present an automated pipeline for low-light VRU labeling. We use LWIR-RGB image pairs and source our labels from LWIR as VRUs differentiate from the background due to their body temperature. Unlike our previous work that relied on a complex hardware setup to align the two modalities, we perform this experiment using a LWIR-RGB stereo setup, lowering hardware complexity and cost, making the pipeline easier to reproduce. We evaluate method end-to-end on a dataset collected in the capital region of Finland. Models trained on our generated labels beat both off-the-shelf baselines and models fine tuned on a hand labeled subset on F1, mAP@50 and LAMR metrics when tested on a manually labeled dataset, and the result holds for two different LWIR detectors. This shows that low-light VRU autolabeling can be made both accessible and effective without specialized alignment hardware. Testing data and code will be made available after acceptance

\end{abstract}


\section{Introduction}

Vulnerable Road User (VRU) detection is a key component in the safe navigation for autonomous vehicles (AVs) and of Advanced Driver Assistance Systems (ADAS) ability to prevent collisions. This task is especially important at night, which carries a higher fatality risk compared to daytime, as reported in \cite{75_25_NIGHT_PED_FATAL}, roughly 75\% of VRU fatalities occur in darkness. This increased risk can be mitigated by improving the detection capabilities of ADAS under low-light conditions.

In addition to the previous causes, VRU detection is also motivated by legislation. Under EU Regulation 2019/2144 \cite{EU2019_2144}, advanced emergency breaking systems capable of detecting vulnerable road users became mandatory for all new passenger cars and light commercial vehicles (categories M1 and N1) in the EU from the 7\textsuperscript{th} of July 2026.

\begin{figure}[t]
    \centering
    \includegraphics[width=0.42\textwidth]{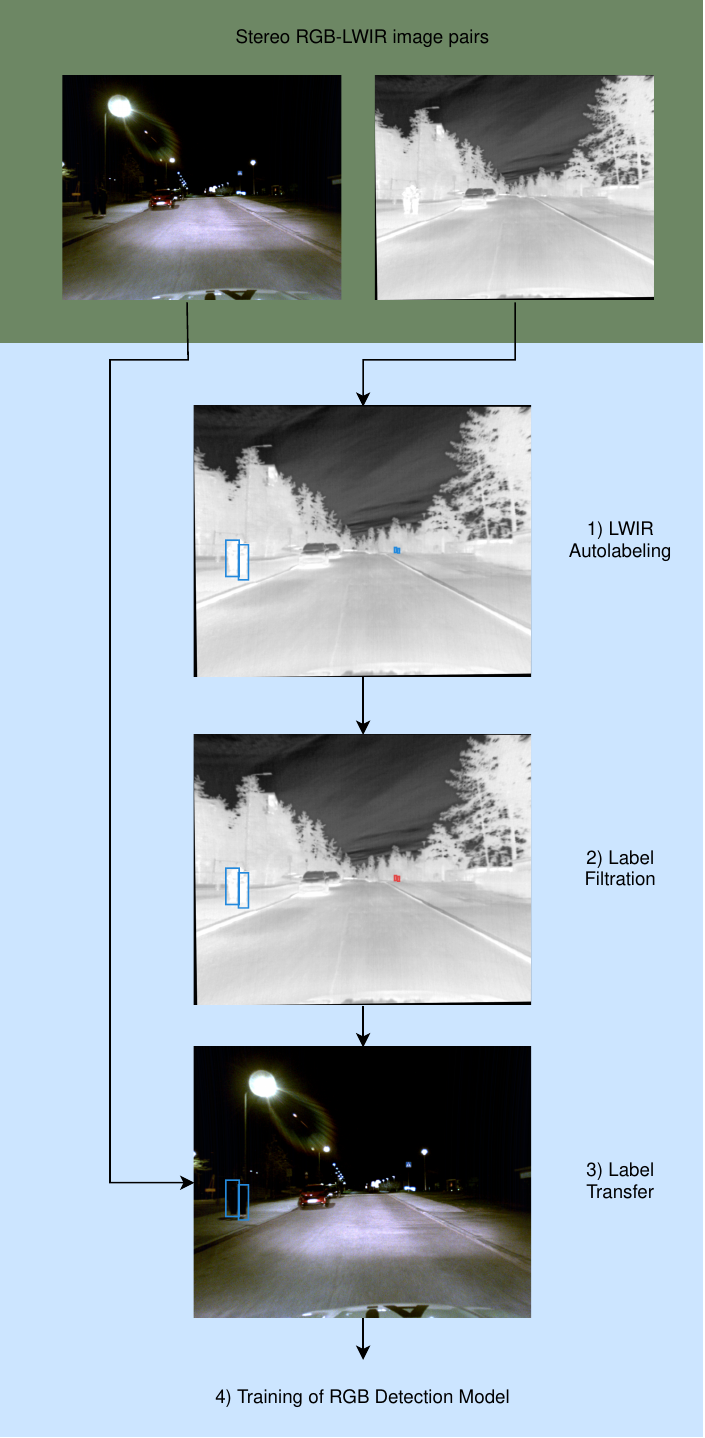}
    \caption{The proposed LWIR-RGB autolabeling pipeline.}
    \label{system-diagram}
\end{figure}


Vehicles currently on the roads face limitations due to the lack of light-invariant onboard sensors, as most are equipped only with RGB cameras. In low-light conditions RGB-based VRU detection is limited due to the lack of adequate illumination. A contributing factor to this issue, is the lack of large-scale low-light datasets \cite{LLVIP}, as labeling overhead is a key limitation in dataset generation. Together, these factors reinforce the need for efficient large-scale autolabeling. 

In this research we propose a novel automated labeling pipeline using a stereo LWIR-RGB setup. Our pipeline exploits the low visible light sensitivity of the LWIR camera. To achieve autolabeling we use overleaping LWIR-RGB image pairs to automatically label VRUs in low-light conditions. The proposed pipeline works by using collected low-light LWIR-RGB image pairs, the LWIR portion of the data is automatically labeled using fine tuned detection models, then filtering and transferring the labels to the RGB portion of the data. This pipeline builds upon our previous work \cite{bouzoulas2025automatic} and aims to deliver a simpler hardware setup method making it more easily accessible. Our experiments were carried out on a dataset recorded in the capital region of Finland in the cities of Espoo and Helsinki. With our main contributions being:

\begin{itemize}
    \item A first of its kind LWIR-RGB stereo autolabeling pipeline for VRU detection 
    \item We prove that our autolabeling pipeline improves VRU detection performance against baseline models and models fine tuned on subsets of the RGB data collected. 
    \item We present an open source dataset for nighttime VRU detection 
\end{itemize}

\section{Related Work}

\subsection{Pedestrian Detection}
 
Pedestrian detection has evolved throughout the years, starting out with hand crafted feature detection, and continuing in today's deep learning era \cite{hand_crafred_feat_to_cnn}. 
An example of early human detection appears in \cite{dalal2005histograms} edge and gradient descriptions are paired with a support vector machine to perform human detection. 
In the coming years with the introduction of deep learning models, convolutional neural networks (CNNs) took over the object detection field through architectures such as R-CNN \cite{rcnn_citation} that learn features directly from data rather than the usage of hand-crafted features. More recently with the emergence of transformers, transformer-based architectures such as DETR \cite{DETR} have matched and often surpassed popular CNN architectures.

In addition to more general object detection models, research efforts have been put towards the development of targeted models specifically for the VRU and pedestrian detection task. On the CNN side a common approach revolves around the advancement of the YOLO \cite{Yolo_original} model family with YOLO-APD being an example of that, where a baseline YOLOv8 was enhanced with attention and feature-fusion modules, 
reaching 77.7\% mAP@50-95 and outperforming the baseline YOLOv8 on a simulated autonomous-driving dataset. Transformer models have followed a similar trend by being adapted to the pedestrian detection task. Such an adaptation was performed on the DETR model family with RC-DETR \cite{crowded_detr} which was developed for pedestrian detection in crowds and is currently one of the benchmarks for the task by achieving a state-of-the-art 38.9\% log-average miss rate (LAMR) on the CrowdHuman benchmark.

Recently foundation models have emerged and gain popularity as a general-purpose alternative to task specific detectors. Segment anything models such as Segment Anything Model (SAM) \cite{sam1_kirillov2023segment} and its successor SAM3 \cite{SAM3carion2025} offer capable zero-shot generalization across a variety of objects with people being one of them. Foundation models require no fine tuning and consequently no training data requiring a single text prompt. 
The generalization and ability to perform with no fine tuning comes at a cost which is explored in \cite{sam_vs_yolo}, where YOLO reached a 72\% F1 score while SAM achieved only 59.8\% on the 
MinneApple dataset \cite{apple_dataset}. In this paper we also explore foundation model by evaluating SAM3's ability to generate labels.

\subsection{Detection in low-light conditions}

Low-light pedestrian detection faces visibility issues as traditional RGB-based detection is highly affected by lighting conditions. 
In spite of that, research efforts have been made in improving low-light RGB pedestrian detection. In \cite{rgb_ther_teacher_student} a fusion method was used for training where both thermal and RGB images were used to distill an RGB only model. Another approach in improving low-light is the artificial brightening of the images, enabling existing models to perform better at task as scene lighting improves and resembles conditions closer to daytime \cite{image_enhancement_for_low_light}. Following a similar principle \cite{light_adaptive_YOLO} appended an image filtering module that was trained in pair with the network, consequently learning which feature enhancements result in the best detections.

To further improve pedestrian detection, many research groups turned towards fusion methods, aiming to reach higher detection accuracy through the use of multiple sensors. Sensor fusion methods have additional complications compared to RGB-based detection methods as they require the purchase and paired calibration of multiple sensors. Nevertheless, fusion methods remain popular as they allow the leveraging of sensor that are less affected by lighting such as thermal cameras and LiDARs. 

A combination of thermal and RGB is often utilized for pedestrian detection tasks as pedestrians differentiate from their environment due to their body temperature. Such a sensor combination was used in \cite{ir-rgb-transformer} paired with a transformer network, the results exceeded state-of-the-art performance on mAP50 and mAP50-95 metrics. Another approach leverages LiDAR and RGB data to improve detection accuracy, such as the method proposed in \cite{lidar-rgb-pedet-nuscenes} where a fusion network that leverages semantic and spatial cues to improve the state-of-the-art performance on the nuScenes dataset \cite{caesar2020nuscenes}. Additionally, more complex setups have been proposed by combining three or more sensors, such as the work done in \cite{thermal-rgb-lidar-dataset} where the R-LiViT dataset was introduced for roadside pedestrian detection showing the potential of trimodal setups.

\subsection{Cross-Modal Labeling and Automated Labeling}

Traditional manual labeling methods are costly in time and as a consequence financially, staggering development. Automated labeling aims to solve that as it removes the need for human effort and speeds up the labeling process \cite{zhang2021survey}. To achieve effective automated labeling results complimentary sensor setups have been used. An example of such a sensor setup is shown in 
\cite{lidar_rgb_autolabelnig}, where automatically generated road segmentations were generated in RGB and then transferred to LiDAR as the goal of the research was to deliver a LiDAR only segmentation network, with the final segmentation network exceeding state-of-the-art method performance on the KITTI dataset \cite{kittidataset}.

Labeling using multiple modalities has also been explored by human labelers. LLVIP \cite{LLVIP} is a thermal and RGB dataset for nighttime pedestrian detection, in it, the authors manually labeled pedestrian instances, but due to the poor RGB visibility of pedestrians, the human labelers used the thermal images to label pedestrians. Following a similar idea our previous work \cite{bouzoulas2025automatic} experimented on the KAIST \cite{KAIST_DATASET} dataset, in our work fine tuned models for IR pedestrian detection were used to label the IR portion of the KAIST dataset, then the detections were transferred to the RGB portion of the dataset, for the label quality evaluation models were trained on ground truth labels and our automatically generated labels, in most cases our generated labels came close or exceeded ground truth label performance when compared on the same human labeled sequence.

\subsection{Research Gap}

Current methods in the literature have approached autolabeling through various angles with different levels of success. Primarily, autolabelers have been exploiting RGB as the label source due to the amount of data available and detector accuracy such as the LiDAR to RGB autolabeling pipeline mentioned in \cite{lidar_rgb_autolabelnig}. Although capable of such methods, they fail to provide the same results in conditions that do not favor RGB detection, such as low-light conditions. 

To address the gap and the need for low-light pedestrian detection datasets we previously proposed an fully automated IR-RGB labeling pipeline. The previously proposed pipeline was successful in proving that generated labels can perform comparatively with human labels, yet a limitation of the method came due to the complex data gathering system it required. In this paper we aim to further narrow the research gap and make low-light pedestrian autolabeling more widely accessible. We do this by using a stereo setup that lowers the hardware complexity, and furthermore to evaluate the feasibility of the method we performed the entire process including stereo camera calibration, extensive data collection and the implementation of the autolabeling pipeline itself resulting in the complete evaluation of the proposed method.

\section{Methods}

\subsection{Pipeline Architecture}
\label{sec:architecture}

Our proposed pipeline consists of three steps. The first is an LWIR labeling step described in section \ref{sec:IR-models}. The second step is a filtration of the generated labels based on their confidence and height relative to the image, and is analyzed in section \ref{sec:label_filtration}. The third step is a label transfer step from the LWIR source to the target RGB images described in section \ref{sec:label_transfer}. The complete overview of the architecture can be seen in Figure \ref{system-diagram}.

\subsubsection{IR detection}
\label{sec:IR-models}

For the LWIR detection task we fine tuned existing models improving their VRU detection capabilities in LWIR images. The fine tuning was performed using a dedicated publicly available dataset. Using the LWIR models we then label the corresponding portion of our dataset. The labels are generated at an extremely low confidence, enabling label filtration.

\subsubsection{Label Filtration}
\label{sec:label_filtration}

In our previous work we explored the process of filtering generated labels based on confidence. In this work we extend this filtration method by adding a label height  parameter to our filtering step.
In our filtration algorithm let $\mathcal{I}$ denote the set of candidate frames and $D_i$ the set of detections transferred from thermal images with $i$ representing the individual frames. Each detection $d\in D_i$ has a confidence value $c_d\in [0,1]$ and a bounding-box height $h_d\in [0,1]$ which is expressed relative to the image height.
The selection is performed based on a confidence threshold $c_t$, a bounding box height threshold $h_t$, a background confidence threshold $c_{bg} << c_t$ and a background sampling ratio $p$. 

\begin{align}
A_i &= \{\, d \in D_i \;:\; c_d \geq c_t  \;\wedge\; h_{d} \geq h_{t\min} \,\}, \label{eq:admissible}\\
R_i &= \{\, d \in D_i \;:\; c_d \geq c_t \;\wedge\; h_{d} < h_{t\min} \,\}, \label{eq:unresolvable}\\
N_i &= \{\, d \in D_i \;:\; c_d \geq c_{bg} \,\}. \label{eq:credible}
\end{align}

Set $A_i$ contains detections that were detected with a high enough confidence and a high enough bounding box to be considered localizable. $R_i$ has detections that are bellow the height floor $h_t$ but pass the confidence threshold $c_t$. Finally, $N_i$ has detections that are above the background confidence threshold, with no height evaluation. Resulting in a set that indicates which frames cannot be considered empty.

Our filtration method uses two filtration parameters $c_t$ and $h_t$. The confidence threshold represents whether a detection can be trusted. The height represents whether or not a detection contains enough visual information to improve models performance and if it can be transferred reliably from LWIR to RGB. 

Using the groups $A_i, R_i, N_i$ we assign each frame a class:

\begin{equation}
\label{eq:frame-class}
\mathcal{C}_i =
\begin{cases}
\textsc{positive}   & \text{if } R_i = \emptyset \;\wedge\; A_i \neq \emptyset,\\[2pt]
\textsc{background} & \text{if } N_i = \emptyset,\\[2pt]
\textsc{discard}    & \text{otherwise,}
\end{cases}
\end{equation}
%
%

In the defined sets POSITIVE frames contain only frames with labels that pass both height and confidence thresholds and no labels that would invalidate the image. BACKGROUND contains only frames where the auto-labeler did not find any detections above the lowest specified threshold $c_{bg}$. Finally, DISCARD contains contains images where the labels did not pass the height-confidence thresholds, yet  they exceeded the $c_{bg}$ value as well as images where labels passed the confidence threshold $c_t$ but did not pass the height threshold $h_t$.




\subsubsection{Label transfer}
\label{sec:label_transfer}

The label transfer process across domains requires the label to be transferred from the source to the target domain while preserving the information it provided in the source. In our case this means that the labels of pedestrians in LWIR need to be transferred to our target RGB images without being shifted in a way that would invalidate the detection. 

As our setup uses a stereo camera setup and mapping imperfections are minor and we are able to perform direct label transfer. This means that the bounding box coordinates in the LWIR portion of the dataset map directly to the RGB images with minimal shifting resulting in this step being performed with no additional overhead.

\subsubsection{Training and evaluation of low-light RGB detection models}
\label{sec:training}

To evaluate the detection capabilities of the models trained on generated labels we experimented with different label filtration values to evaluate how they affect model performance. Next, to validate and test the models a validation and testing set was manually labeled. For the final tests we compare the models trained on our generated labels against baseline models and models fine tuned on the validation data to provide some domain adaptation.

\subsection{Sensor Setup}

\begin{figure}[tbp]
    \centering
    \includegraphics[width=0.45\textwidth]{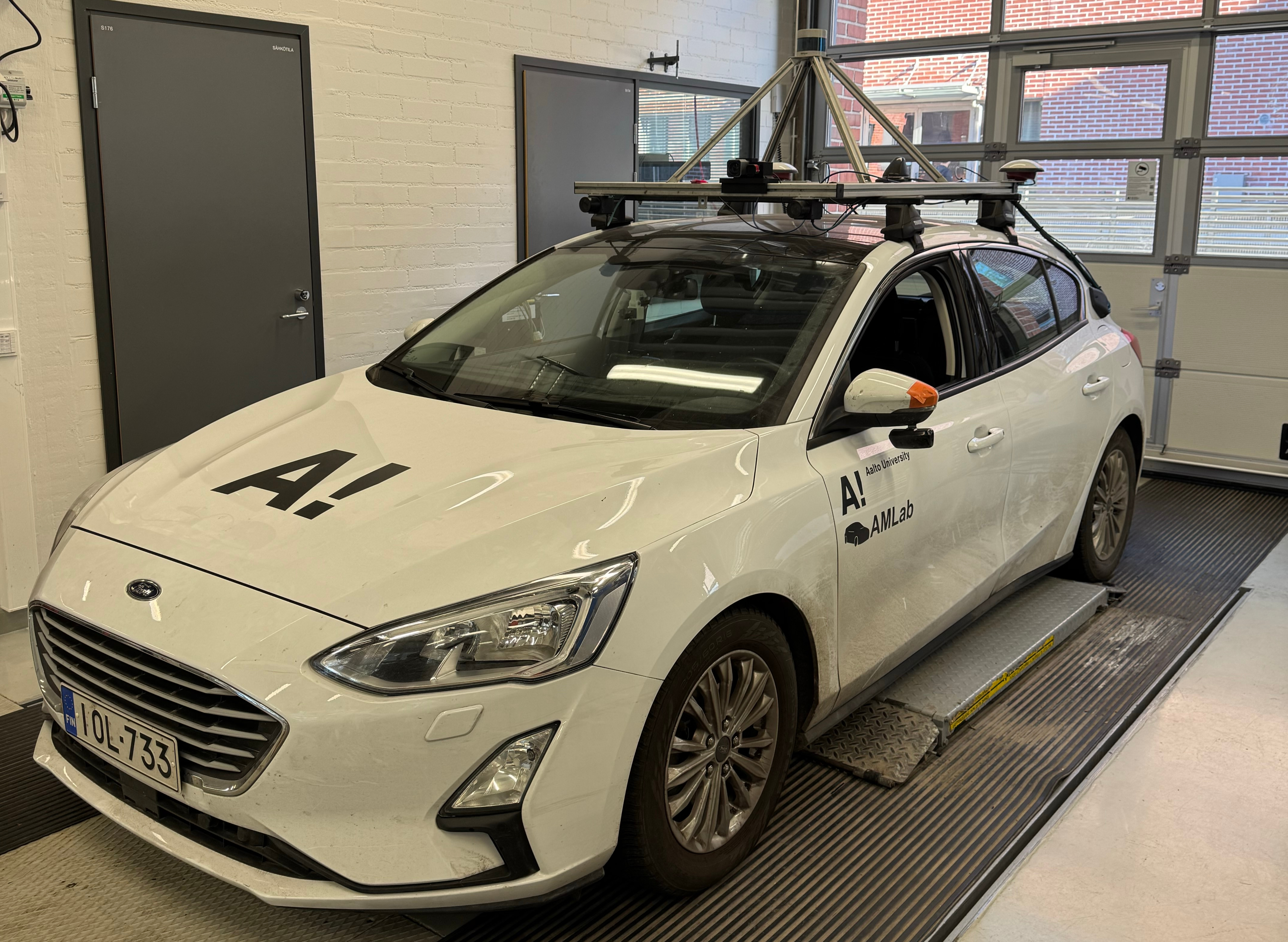}
    \includegraphics[width=0.45\textwidth]{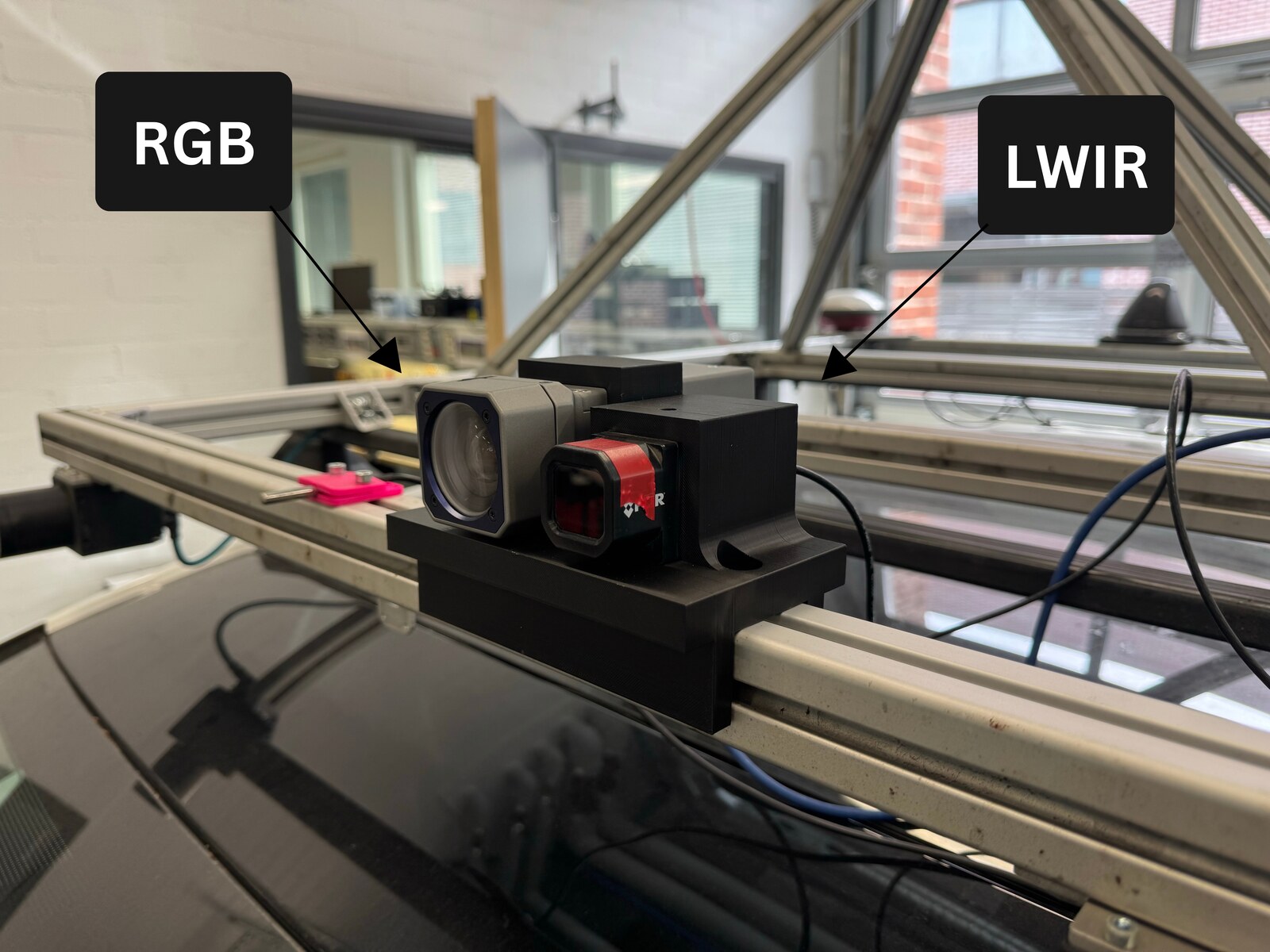}
    \caption{LWIR-RGB stereo camera setup on the research vehicle.}
    \label{camera_setup_picture}
\end{figure}

\subsubsection{Hardware and Mounting}

To achieve LWIR-RGB automated labeling, overlapping image pairs are required. To achieve the desired overlap an LWIR-RGB stereo setup was built. Our setup consisted of two cameras, one LWIR and one RGB. The camera models used were FLIR Blackfly S BFS-PGE-13Y3C-C for RGB and FLIR-Boson-GE-640 LWIR. Additionally, the RGB camera was equiped with a  wide field of view (FoV) lense (4mm UC Series Fixed Focal Length Lens) resulting in a horizontal FoV of $\ang{84.8}$, and the default FoV of the thermal camera being $\ang{75.0}$. The wider lens was chosen to allow the full utilization of the thermal camera data. 

For our experiments the sensors were positioned on the car's roof. The placement was chosen as the thermal camera cannot be positioned behind the windshield due to the glasses ability to block LWIR waves. The cameras were secured next to each other and aligned. Finally, to avoid damage from the weather the RGB camera was placed in a water-proof casing 
The sensor placement on the research vehicle is shown in Figure \ref{camera_setup_picture}

\subsubsection{Temporal Camera Synchronization}

The synchronization of the system was performed using Precision Time Protocol (PTP). The hardware selected faced limitations as the LWIR camera did not provide \textit{fire on command} capabilities that would enable both cameras to snap images at the same instant with minimal delays. To overcome this issue, the cameras were set to record data freely at maximum frequency and store it alongside the PTP timestamp of the instant the image was captured. After the data collection process had concluded the images were paired using a sliding window of 10$ms$, resulting in a maximum delay between an image pair of 10 milliseconds, during the pairing process each image was used at most once if an available pair was found in the 10$ms$ time window.

\subsubsection{Stereo Calibration} 

\begin{figure}[htbp]
    \centering
    \captionsetup[subfigure]{position=top,justification=centering}

    \begin{subfigure}{0.48\linewidth}
        \caption{Raw LWIR}
        \includegraphics[width=\linewidth,keepaspectratio]{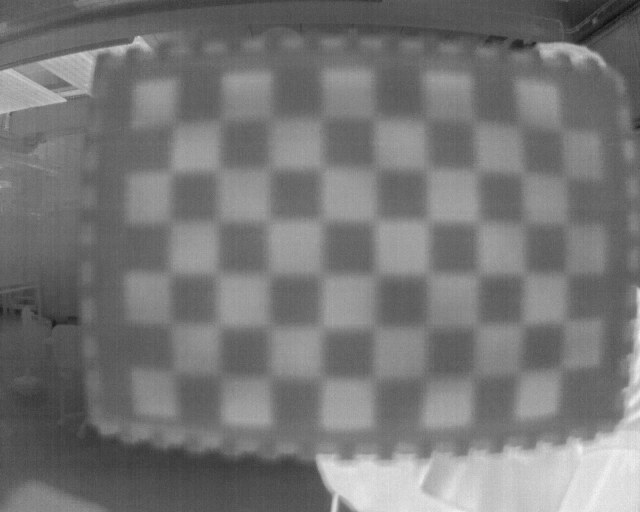}
    \end{subfigure}
    \hfill
    \begin{subfigure}{0.48\linewidth}
        \caption{Raw RGB}
        \includegraphics[width=\linewidth,keepaspectratio]{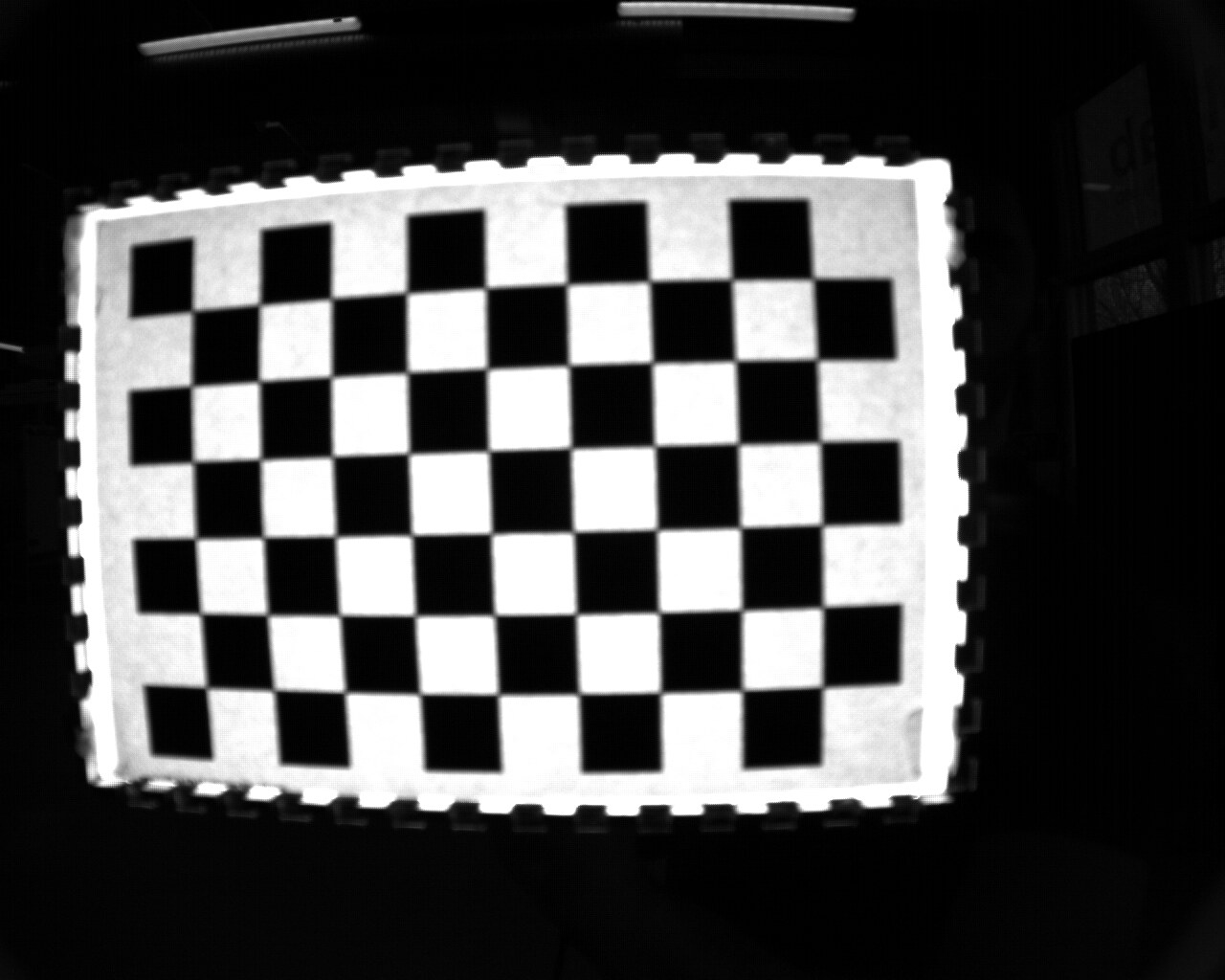}
    \end{subfigure}
    \begin{subfigure}{0.48\linewidth}
        \caption{Undistorted LWIR}
        \includegraphics[width=\linewidth,keepaspectratio]{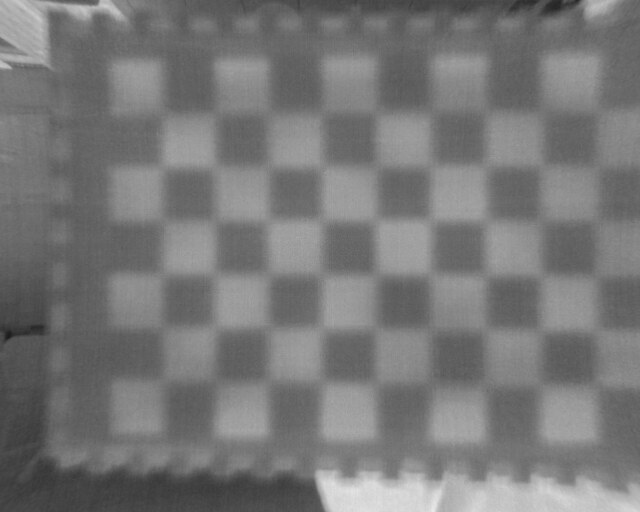}
    \end{subfigure}
    \hfill
    \begin{subfigure}{0.48\linewidth}
        \caption{Undistorted RGB}
        \includegraphics[width=\linewidth,keepaspectratio]{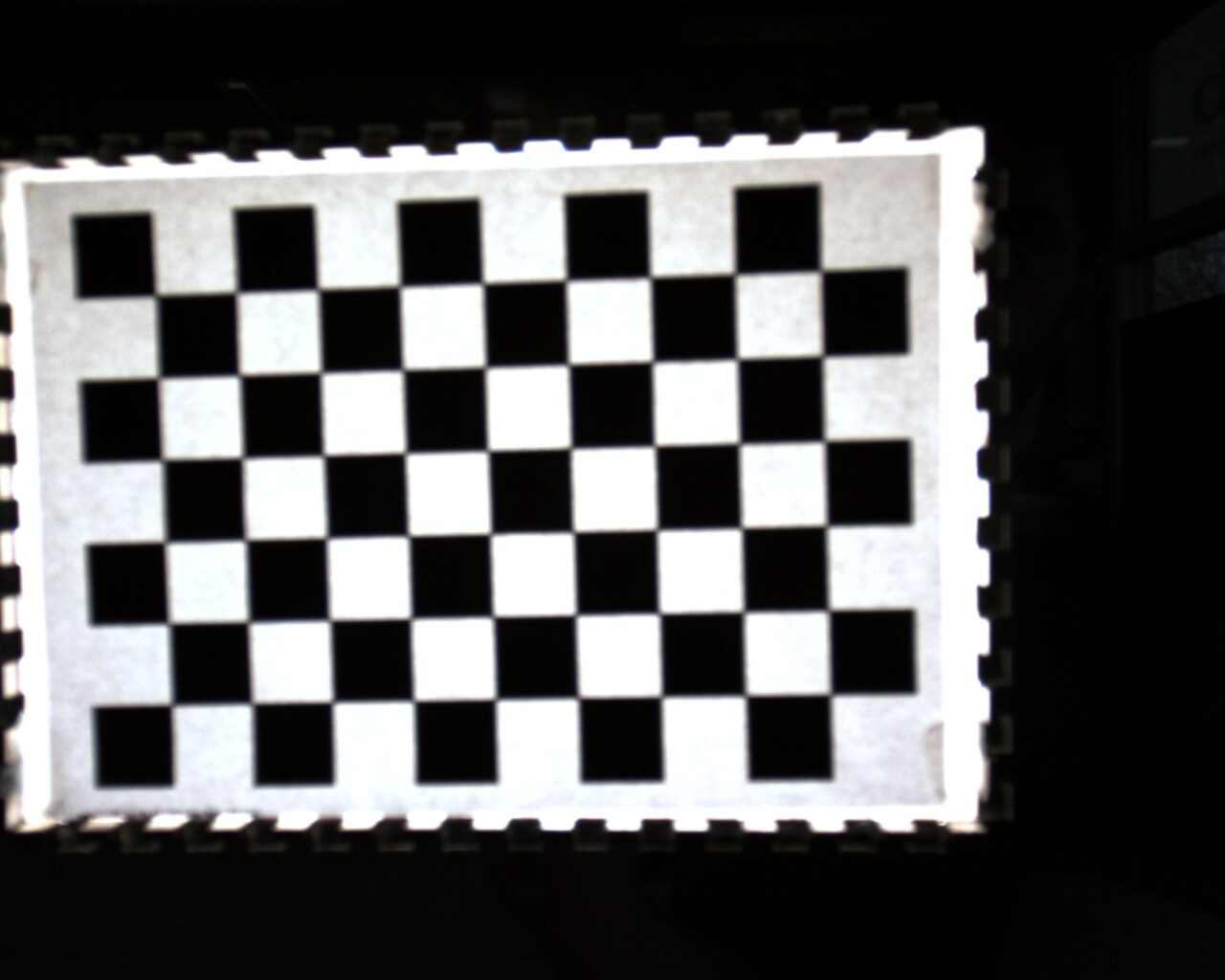}
    \end{subfigure}

    \caption{LWIR and RGB chessboard images captured during calibration, with the chessboard being illuminated with a flood light. Top row corresponds to original images, and bottom row shows images undistorted with the calibrated parameters.}
    \label{chessboard_distorted_undistorted_comp}
\end{figure}

The stereo calibration of an LWIR-RGB setup can be performed in two primary ways. The main difference concerns the use of a target, resulting in target-based and target-less calibration. Target-less requires no target and is done online meaning that for each LWIR-RGB image pair modality invariant features are derived and the calibration is performed based on them \cite{yang2024autonomous}. For target based methods a target is used, often in the form of a chessboard such as the one we used for our calibration setup. More specifically we calibrate using an active target \cite{active_passive_calibration} method by combining a paper pattern and placing a floodlight behind it which acts as the heat source. Our pattern had $9 \times 6$ squares and the squares had sides of $25 mm$. Using the setup 57 images were collected for the stereo calibration. The calibration resulted in the root mean square (RMS) projection error of 1.315, RGB error of 0.407 and LWIR error of 0.775. Images showcasing the distorted and undistorted calibration board can be seen in Figure \ref{chessboard_distorted_undistorted_comp}.

\begin{figure*}[t]
    \centering
    \begin{tabular}{GGG}
        \textbf{RGB} & \textbf{LWIR} & \textbf{Calibration} \\
        \midrule
        \gridimg{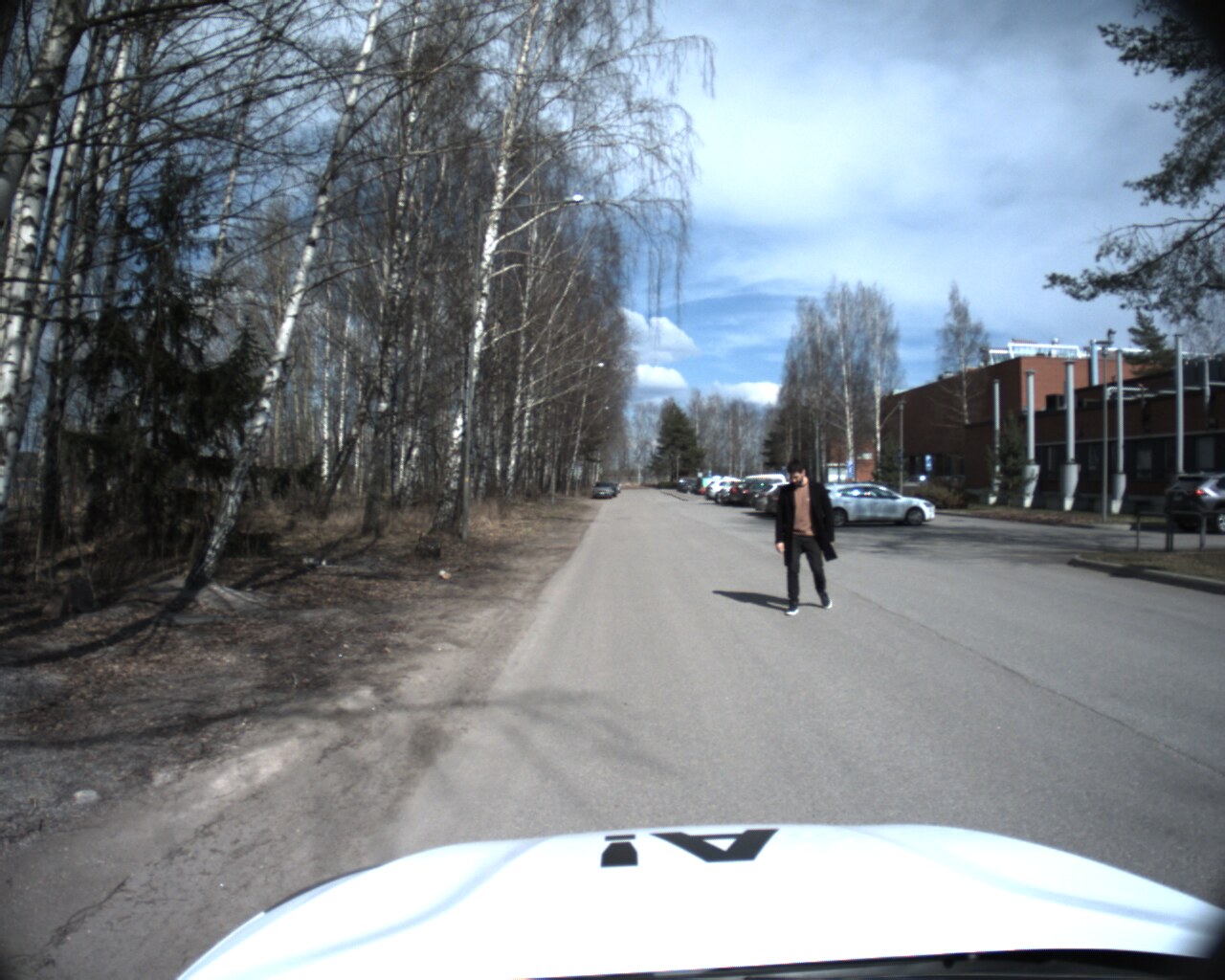} &
        \gridimg{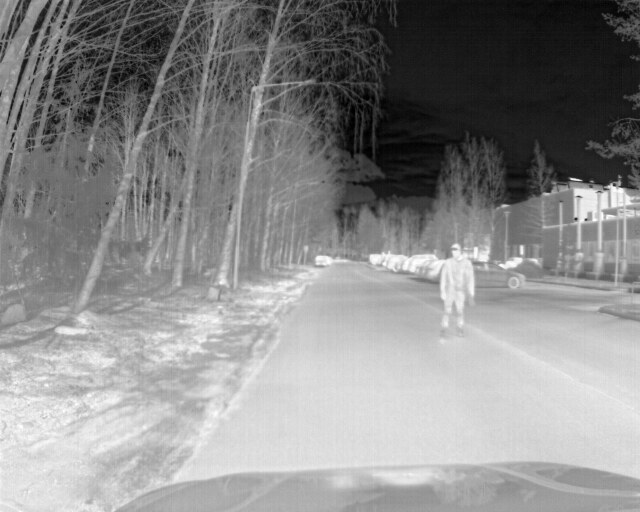} &
        \gridimg{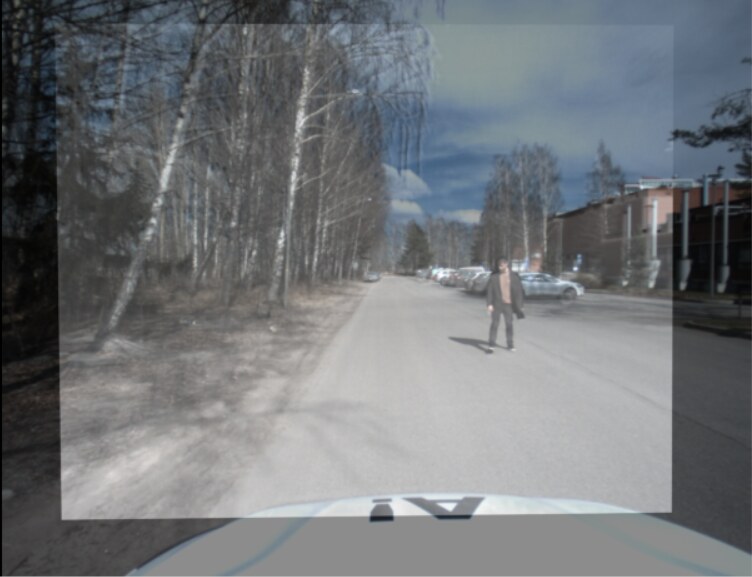} \\
        \addlinespace[4pt]
        \gridimg{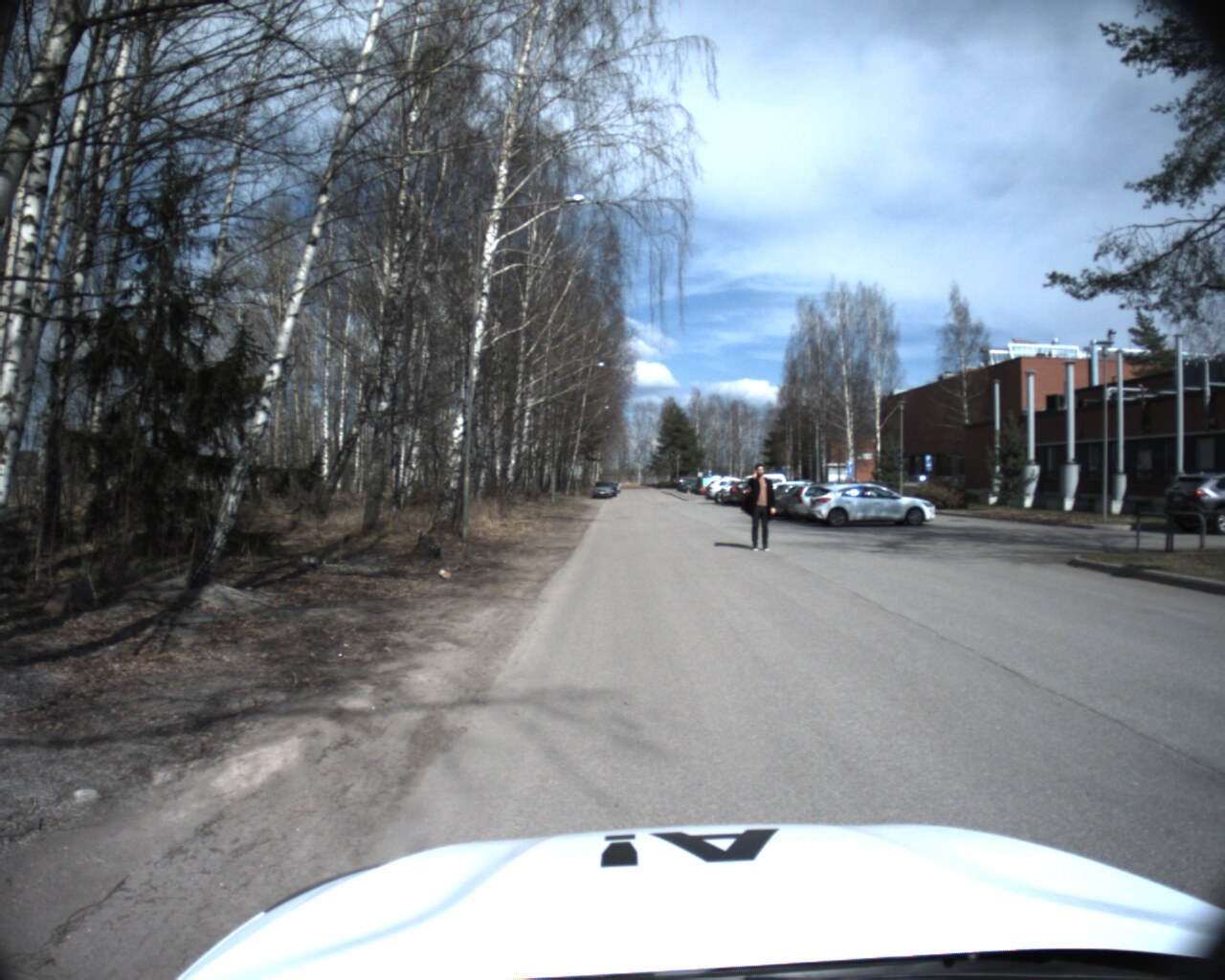} &
        \gridimg{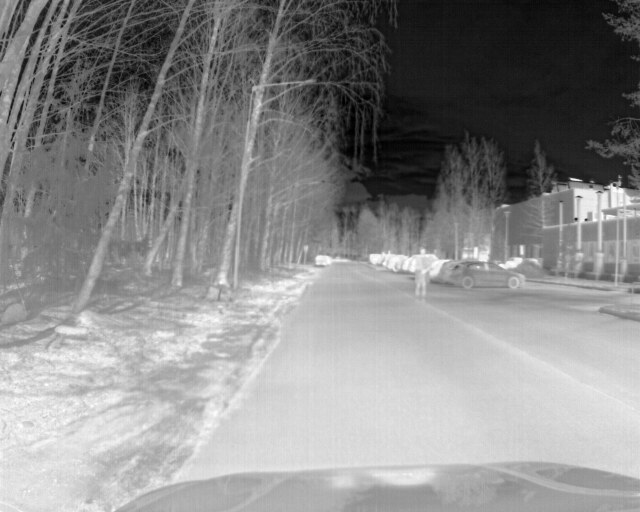} &
        \gridimg{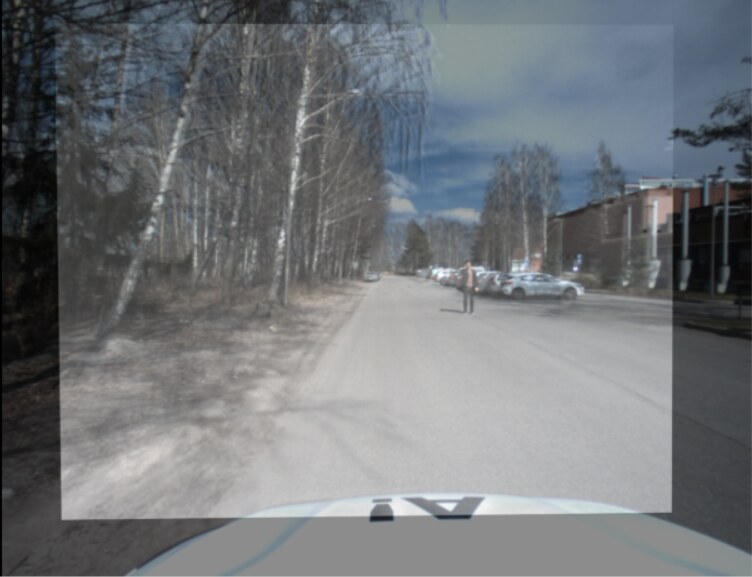} \\
        \addlinespace[4pt]
        \gridimg{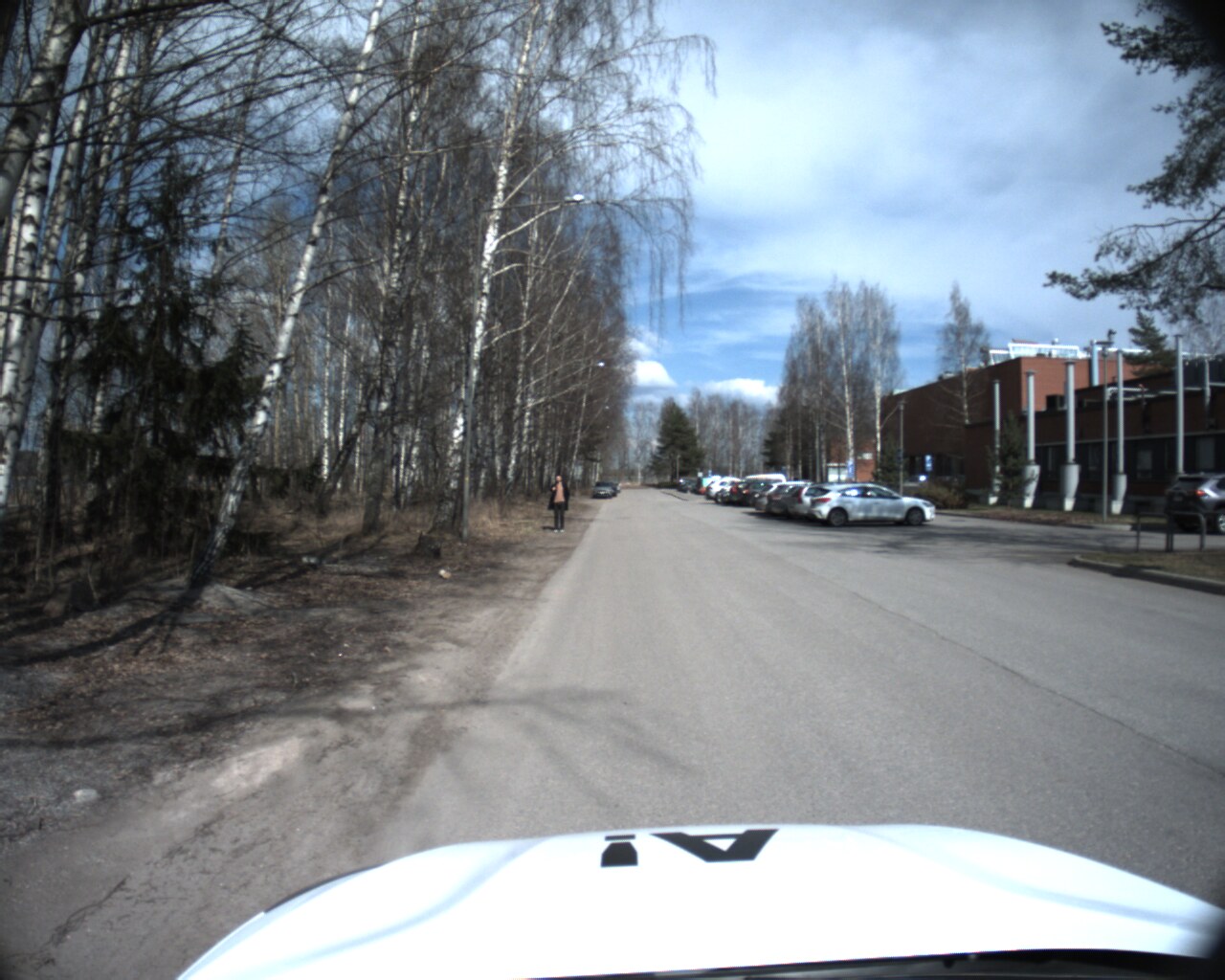} &
        \gridimg{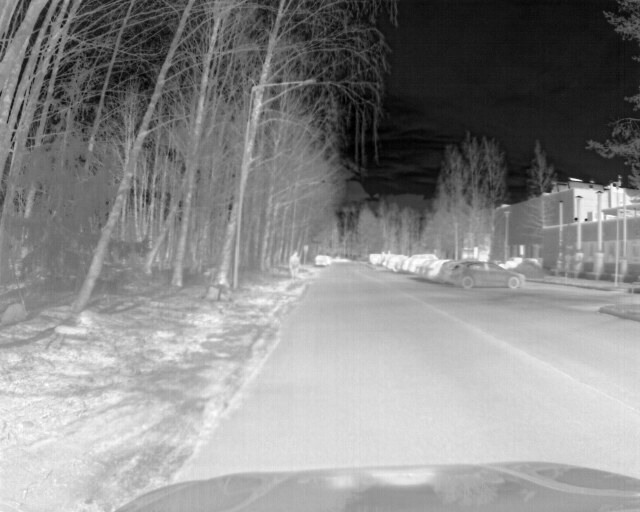} &
        \gridimg{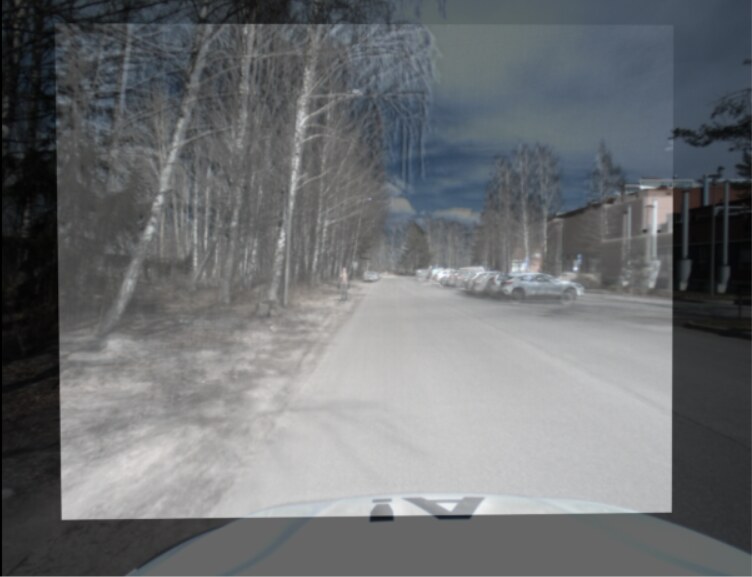} \\
        \addlinespace[4pt]
        \end{tabular}
    \caption{Visualizations of corresponding RGB and LWIR images, with the rectified images overlaid on top of each other to show the quality of the calibration.}
    \label{manually_calibrated_stereo}
\end{figure*}

The calibration was further tested using a pedestrian as a reference point in multiple locations and distances. During the tests, it was noted that a minor error on the vertical axis occurred after the calibration, which did not improve regardless of the number of re-calibration attempts, the mentioned shift appeared on the vertical axis and was corrected with a manual update of the translation matrix of the calibration file, the individual RGB and LWIR images alongside the calibration are shown in Figure \ref{manually_calibrated_stereo} for three pedestrian locations.

Additionally, the tests allowed for the evaluation of the calibration at several distances. As the setup has the two cameras slightly apart the object appears in a slightly different position in the two images based on the distance from the vehicle. As a result the calibration aligns the two images precisely only at one distance with differences appearing the further away from that point an object is. The best overall alignment between the modalities was achieved at $z = 20\,\text{m}$ which was chosen as the convergence distance for the setup


\subsection{Dataset}
\label{sec:data}

\begin{figure*}[t]
    \centering
    \begin{tabular}{GGGG}
        \midrule
        \gridimg{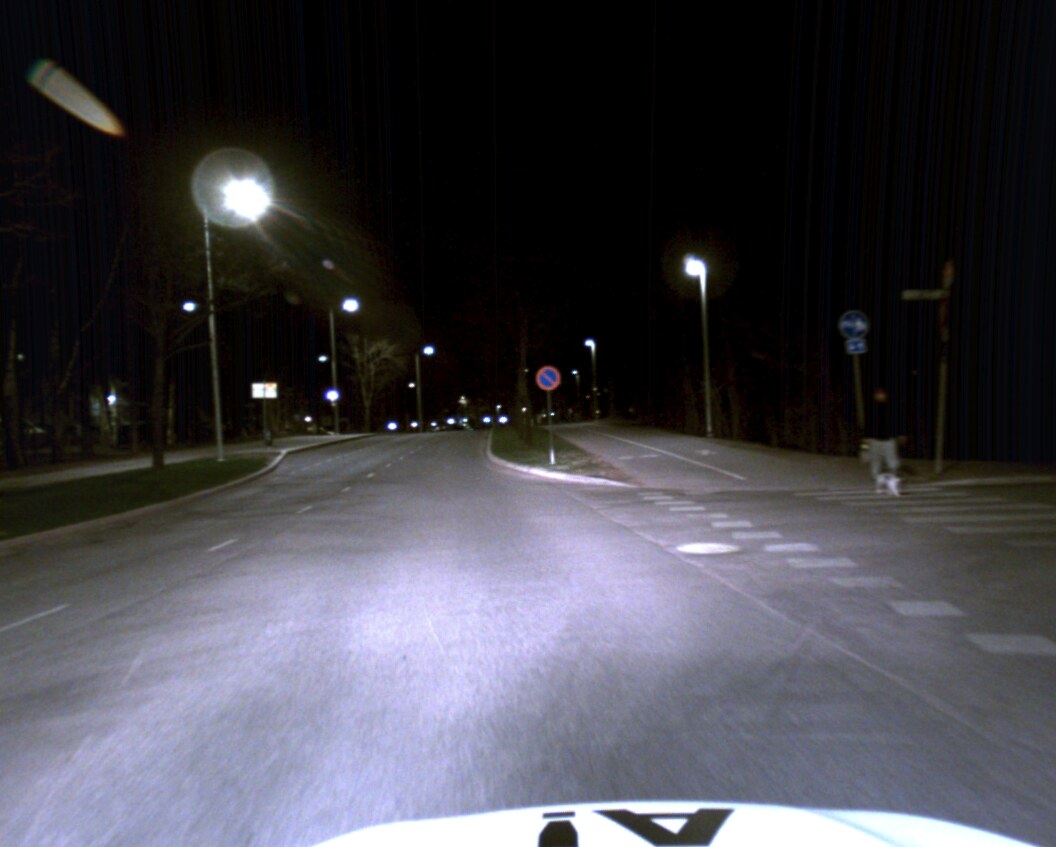} &
        \gridimg{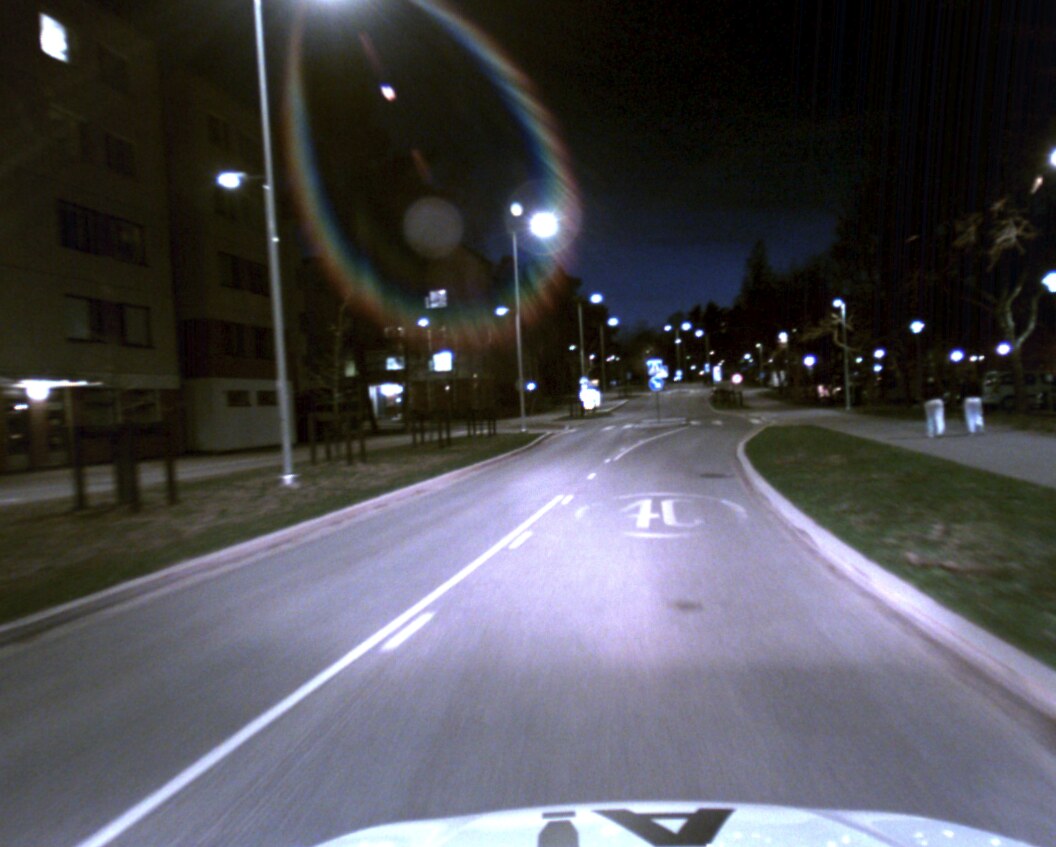} &
        \gridimg{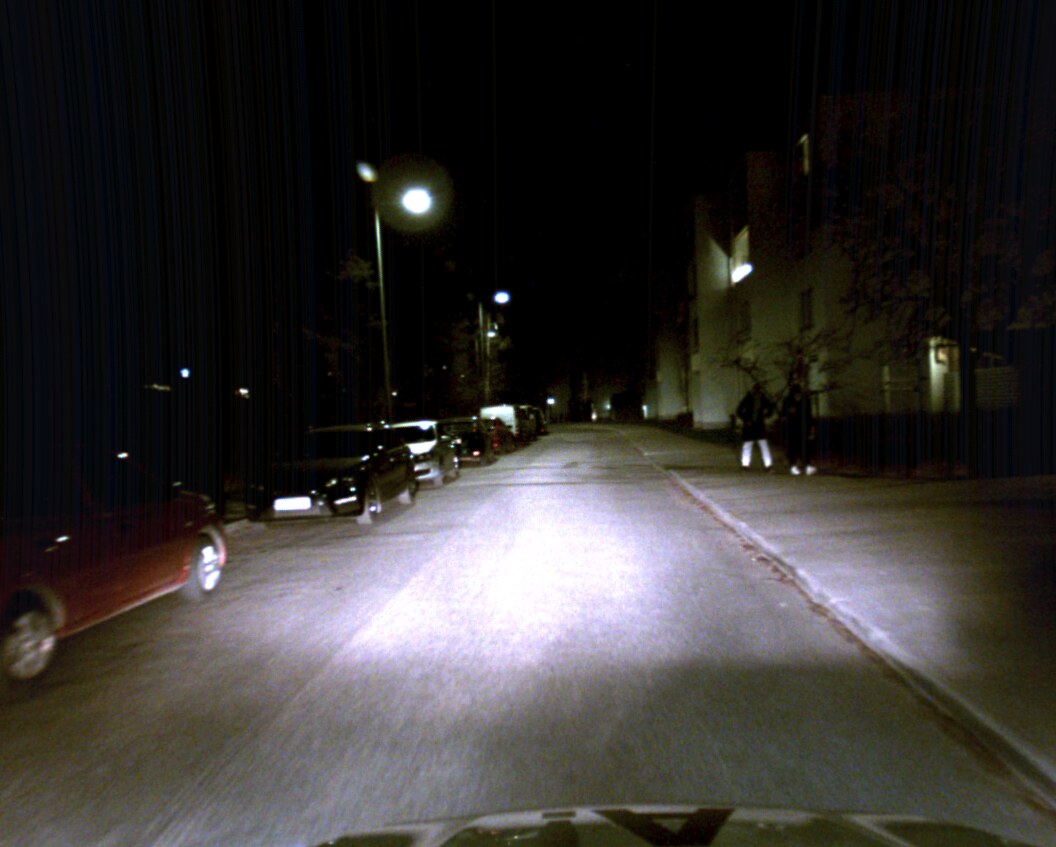} &
        \gridimg{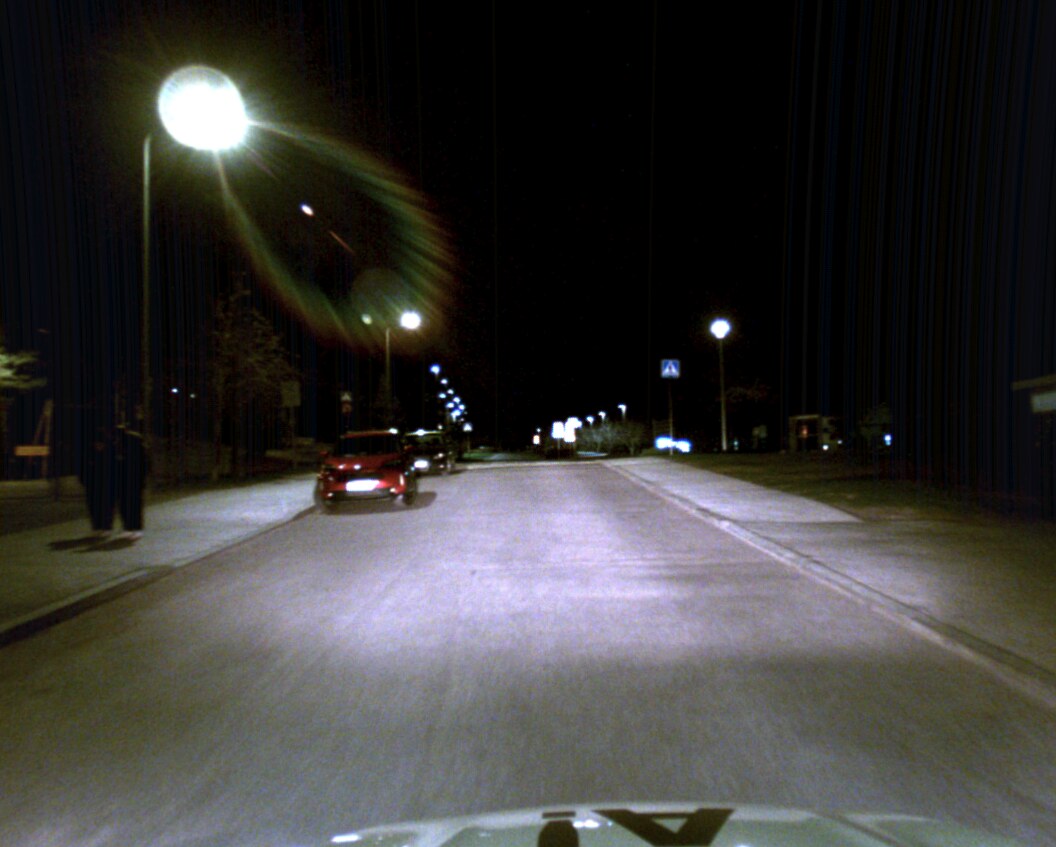} \\
        \addlinespace[4pt]
        \gridimg{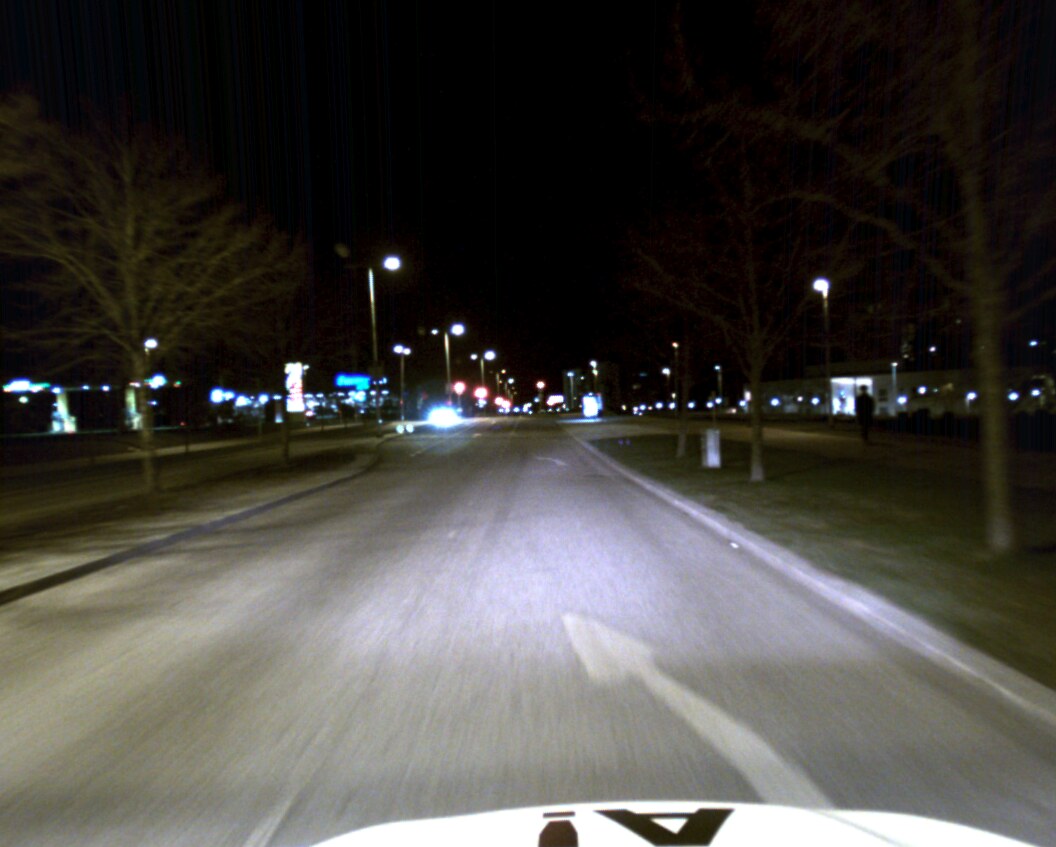} &
        \gridimg{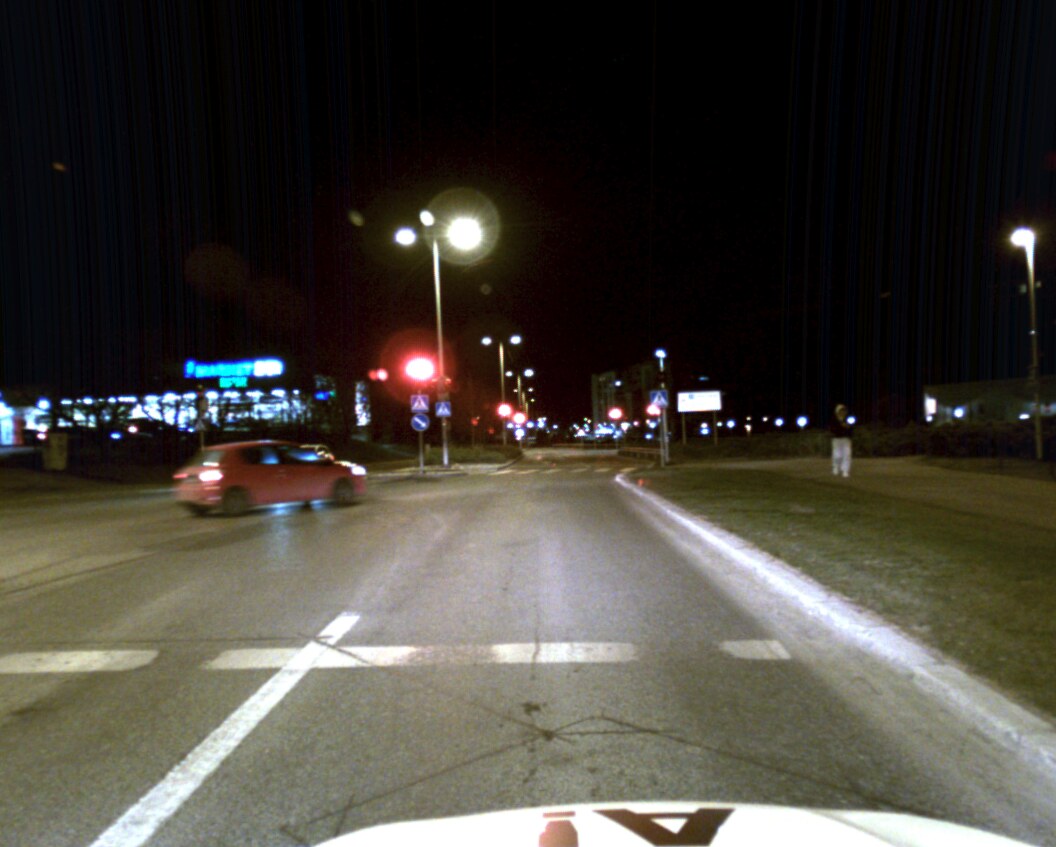} &
        \gridimg{./resources/train_data_samples/espolahti_29_4_26_20260429_224810_rgb_000958.jpg} &
        \gridimg{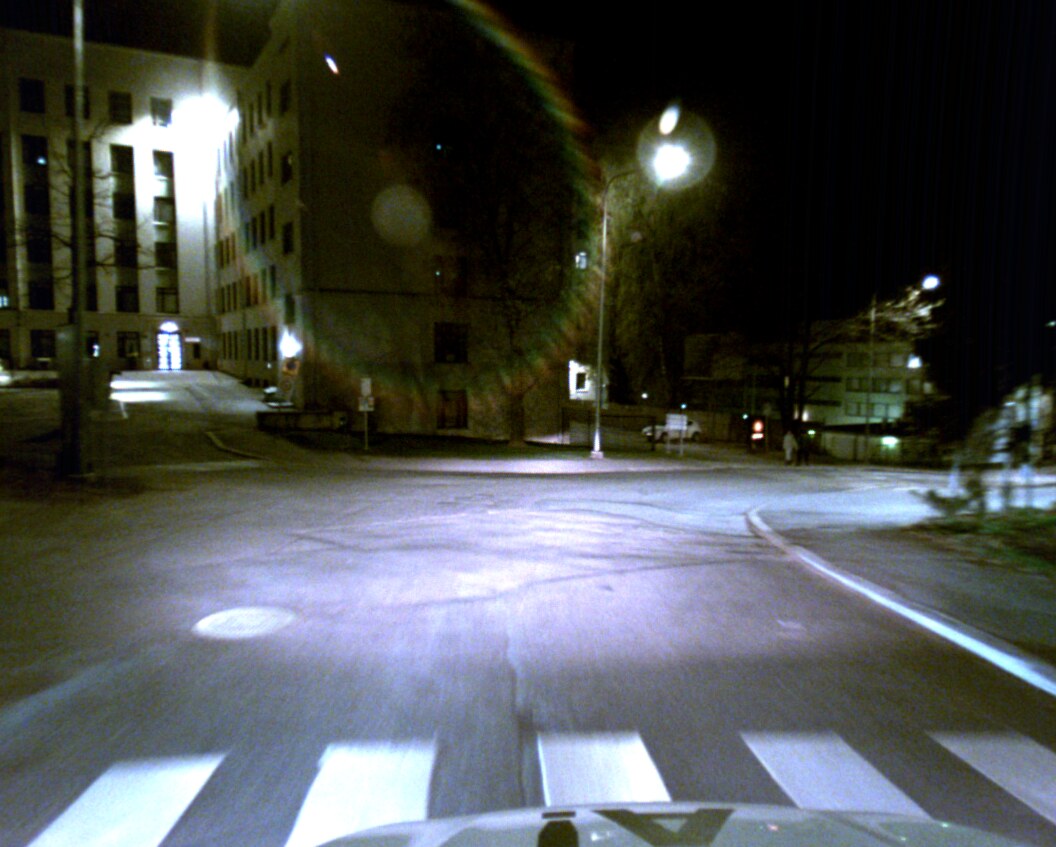} \\
        \addlinespace[4pt]
        \gridimg{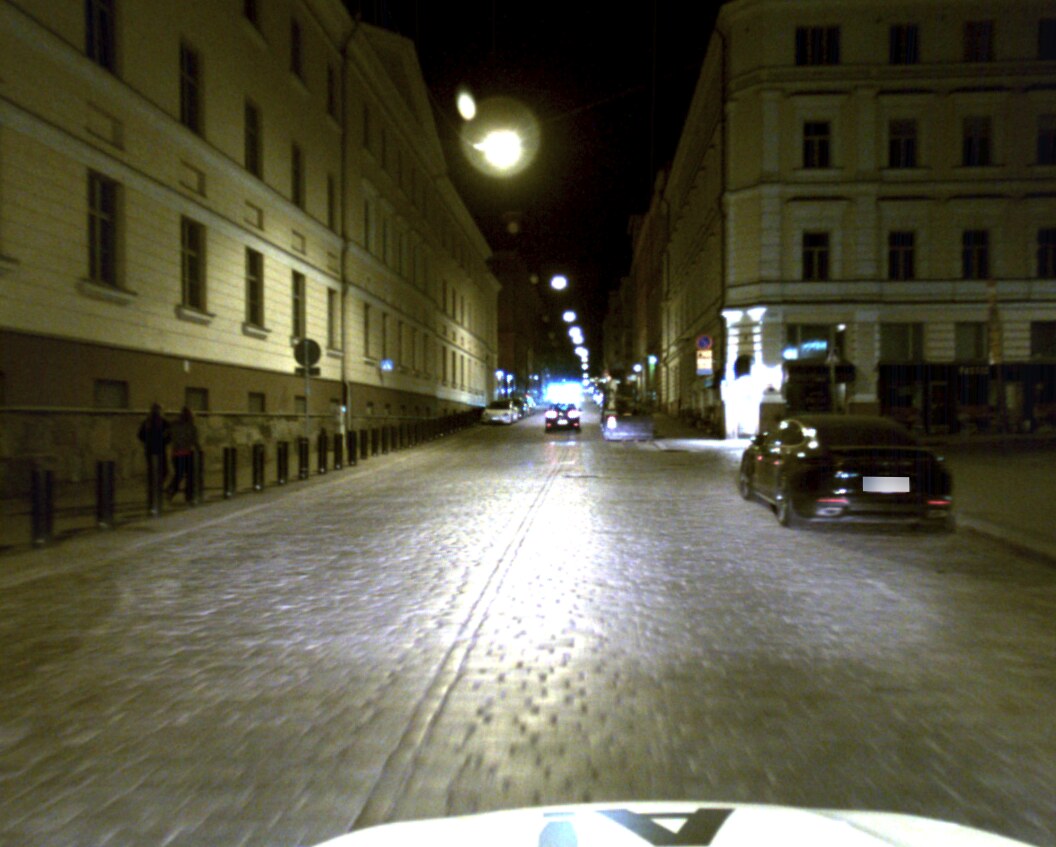} &
        \gridimg{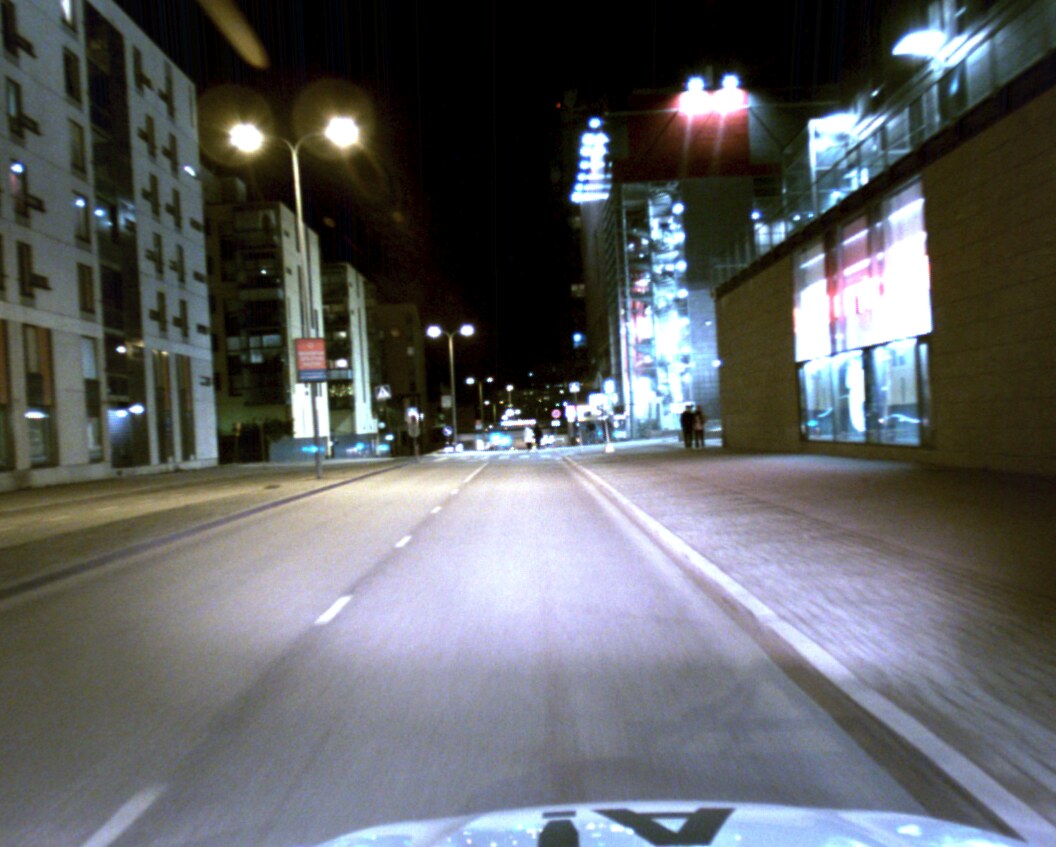} &
        \gridimg{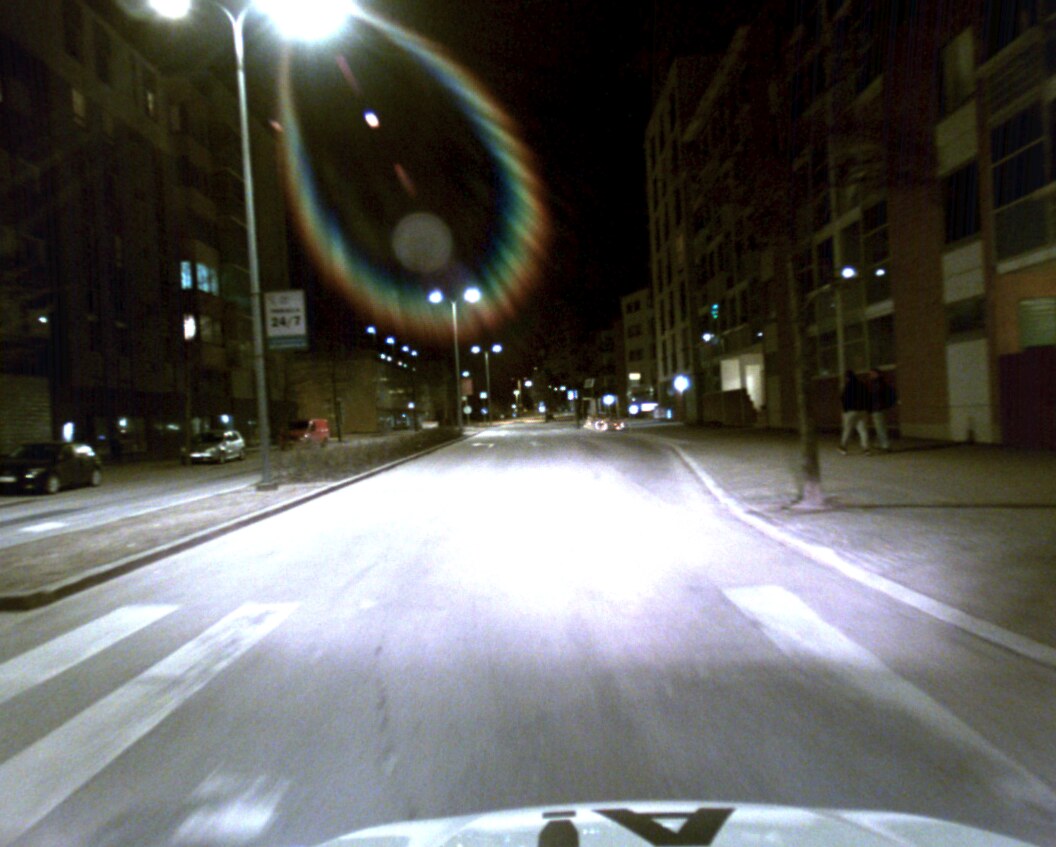} &
        \gridimg{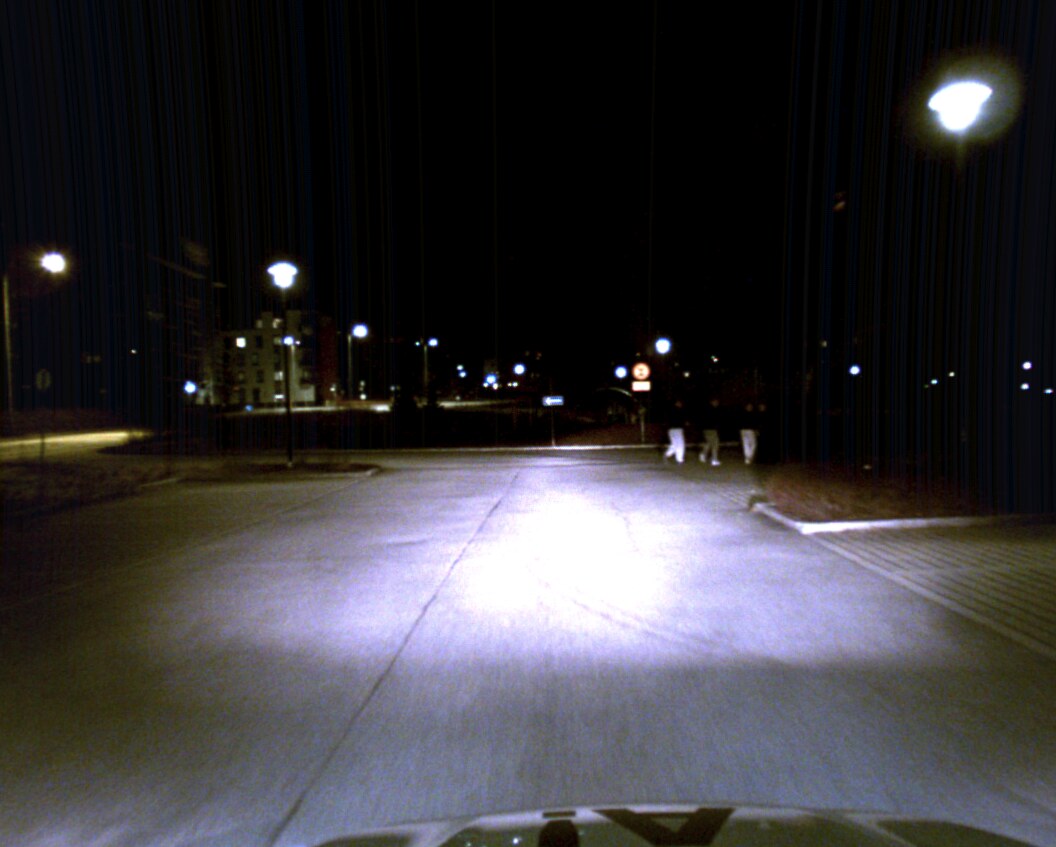} \\
        \addlinespace[4pt]
        \end{tabular}
    \caption{Training data samples collected from various locations during data collection.}
    \label{train_data_exampls}
\end{figure*}

\subsubsection{Data Collection}

For the data collection process the capital region of Finland was chosen due to its geographic proximity. More specifically the data collection covered various locations in the cities of Espoo and Helsinki. The data collection was performed during late winter and early spring over the course of two months, covered a rough geographic area of 56 square kilometers ($km^2$) and took $~60$ hours of driving. The data collection process aimed to include a diverse set of scenes, as a result data was collected in both dense metropolitan and suburban areas. 
The final dataset was aimed to reflect various VRUs in low-light driving conditions. Examples of our collected images can be seen in Figure \ref{train_data_exampls}.

The recorded data was preprocessed before being used to ensure the anonymization of the VRUs, vehicles and other citizens that might appear in the dataset. The anonymization was done using EgoBlur \cite{egoblur_by_meta}, and was done for both faces and license plates. 


\subsubsection{RGB Model Evaluation Data}
For the evaluation process of our final RGB models a subset of the dataset was manually annotated. For this task out of the recorded sequences two were chosen to be used for our validation and testing. The sequences were chosen from an area that does not appear anywhere in the training dataset. The images were subsampled at 1 frame per second to ensure that all VRU instances that appear are unique and a single part of the sequence doesn't bias the results due to its over representation. Our validation data was 441 images with 336 pedestrian instances, and testing set included 1400 images with 1173 instances. 

\subsubsection{IR Model Training Data}
In order to achieve reliable VRU detection in the LWIR images special LWIR detection models were trained.
The models required training data that would closely resemble the recorded data from Helsinki.
To achieve this the FLIR-ADAS \cite{flir_adas_v2_dataset} dataset was chosen.
The choice was made as the dataset was produced by the same company that the camera was purchased from, 
resulting in the minimization of the domain gap.


\section{Results}

\subsection{LWIR Model Performance} 

For our LWIR detectors SAM3, YOLOV11l, and RF-DETR were explored. More specifically we use SAM3 as a foundation model providing it only with a prompt "person" while YOLOv11l and RF-DETR were fine tuned on LWIR data. The fine tuning was performed on the FLIR-ADAS dataset. We used training and testing data for the training combined and only validated on the validation set. This was done with the goal of maximizing the number of samples provided to our LWIR labeling models. The evaluation results of all the models are shown in Table \ref{IR_model_validation_results}, the table has all of the best results bolded with YOLOv11l having the best performance across all metrics, followed by RF-DETR and SAM3 having the poorest. 

\begin{table}[H]
    \centering
    \caption{Evaluation of the LWIR detection models on the FLIR-ADAS dataset. Bolded values indicate the best result in each column.}
    \setlength{\tabcolsep}{2pt} 
    \begin{tabular}{lccccc} 
    \toprule
    \textbf{Base Model} & \textbf{Training Data} & \textbf{P} & \textbf{R} & \textbf{mAP@50} & \textbf{mAP@50-95}  \\
    \midrule 
    RF-DETR  & FLIR-ADAS & 0.682 & 0.5653 &  0.801 & 0.452 \\ 
    YOLOv11l & FLIR-ADAS & \textbf{0.827} & \textbf{0.750} &  \textbf{0.827} & \textbf{0.501} \\
    SAM3     & None & 0.560 & 0.315 &  0.339 & 0.110 \\  
    \bottomrule
    \end{tabular}
    \label{IR_model_validation_results}
\end{table}

\subsection{RGB Detection Model Evaluation} 

For the validation of our pipeline and automatically generated labels we followed a two step process. Initially we narrowed down which pedestrian heights yield the best results and afterwards using these values we experimented on confidence. Additionally throughout our experiments we keep a consistent upper empty image threshold of $p = 10\%$ of the labeled subset, and a background confidence threshold $c_{bg} = 0.02$ these parameters stay fixed throughout our experiments to ensure that our results do not get affected by an excessive amount of empty frames.

For the pedestrian height experiments we fixed the confidence value of our labels and experimented only on the pedestrian height parameter. 
More specifically, we set the confidence to 0.25 for the height experiment as in our previous research \cite{bouzoulas2025automatic} that confidence value resulted in the best performing labels. 
Using the fixed confidence value we generated labels at pedestrians heights starting at 10\% and going down to 0\% with a 2\% step, to evaluate the heights that yield the best results and whether or not the height parameter provides any value at all. The upper limit of 10\% was used as popular benchmarks such as KAIST\cite{KAIST_DATASET} and Caltech\cite{caltech_pedestrian_dataset} have used it as the highest "reasonable" height for detection tasks, as a result, we explore that as the highest threshold as well. Our tests are shown in Table \ref{tab:height_sweep}, and evaluated based on mAP@50. The results showed that overall 4\% and 2\% labels yielded the best results

\begin{table}[tbp]
    \caption{RGB detector model accuracies reported as mAP$_{50}$ on the test sequence, with different height thresholds used in the autolabel filtering. Confidence threshold selected as $c_t = 0.25$. Best result per column in bold, second best underlined. Number of frames corresponds to the amount of image samples remaining in the training data after filtering.}
    \label{tab:height_sweep}
    \centering
    \begin{tabular}{c rcc rcc}
        \toprule
        & \multicolumn{3}{c}{LWIR-YOLO labels} & \multicolumn{3}{c}{LWIR-DETR labels} \\
        \cmidrule(lr){2-4}\cmidrule(lr){5-7}
        $h_t$ & Frames & YOLO & DETR & Frames & YOLO & DETR \\
        \midrule
        0.10 & 6164  & 0.208 & 0.161 & 5163  & 0.165 & 0.161 \\
        0.08 & 9489  & 0.226 & 0.187 & 8123  & 0.195 & 0.190 \\
        0.06 & 16455 & 0.252 & 0.210 & 13914 & 0.238 & 0.207 \\
        0.04 & 27930 & 0.243 & \textbf{0.225} & 24617 & \underline{0.250} & \textbf{0.228} \\
        0.02 & 51483 & \textbf{0.282} & 0.219 & 53684 & \textbf{0.252} & 0.215 \\
        0.00 & 54854 & \underline{0.254} & \underline{0.224} & 65730 & 0.243 & \underline{0.221} \\
        \bottomrule
    \end{tabular}
\end{table}

As 2\% and 4\% labels generated the best results, they were used for the confidence experiments. For our confidence values we chose 0.1, 0.25 and 0.5. The full results of our experiments are shown Table \ref{tab:final_results_full}.

To evaluate how well the models perform we compared them against baselines on our test data. The baselines come in two forms, one being the baseline COCO pretrained YOLO and DETR models as well as SAM3 with the prompt "person", the other baseline to compare against was YOLO and DETR fine tuned on our validation set consisting of 441 images with 336 VRU instances, aimed at evaluating the performance gains achieved by labeling a small subset for domain adaptation.

In our tests the best results were achieved with YOLOv11l trained on LWIR-YOLO labels, the results were achieved using a confidence value of 0.25 and a detection height threshold of 2\%.
On our key evaluation metric mAP@50 the best performing model trained on generated labels beat every baseline including its own baseline by 41.7\% and the best overall baseline (DETR) by 38.2\%. Additionally, the model beat all baseline models across all key metrics, F1, mAP@50, mAP50-95, and LAMR. This result shows how autolabeling can beat a model trained on a manually labeled subset, and high-end baseline models. 

A key result of our experiments is that autolabeling performance is not tied to a single model. This is demonstrated by the results of models trained on LWIR-DETR labels, as they did not yield the best results but closely followed the performance trends of the models trained on LWIR-YOLO labels, consistently exceeding all baselines in the mAP@50, and LAMR metrics. In contrast to the task specific detectors SAM3 showed the worst results, as none of the models trained on its labels exceeded any baseline. This difference can be attributed to SAM3 being a general purpose model, and not a dedicated and fine tuned detection model, this points towards the remaining performance gap between task specific detectors and foundation models. 

\begin{figure*}[tbp] 
    \centering
    \begin{subfigure}[b]{0.44\textwidth}
        \centering
        \includegraphics[width=\linewidth,keepaspectratio]{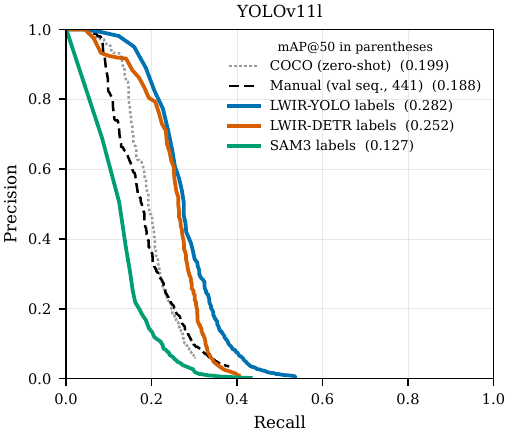}
        \caption{YOLO -- PR}
        \label{fig:yolo_pr}
    \end{subfigure}
\hspace{0.04\textwidth}
    \begin{subfigure}[b]{0.44\textwidth}
        \centering
        \includegraphics[width=\linewidth,keepaspectratio]{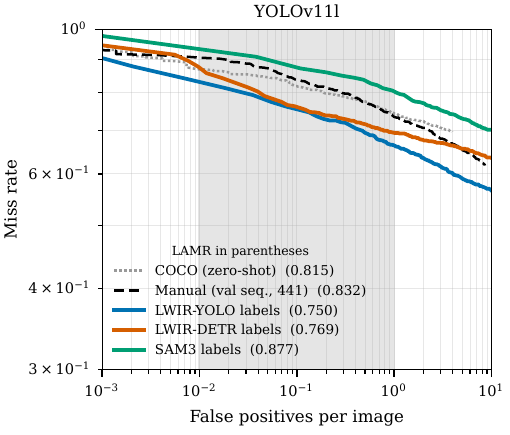}
        \caption{YOLO -- MR}
        \label{fig:yolo_mr}
    \end{subfigure}

    \vspace{0.5em}

    \begin{subfigure}[b]{0.44\textwidth}
        \centering
        \includegraphics[width=\linewidth,keepaspectratio]{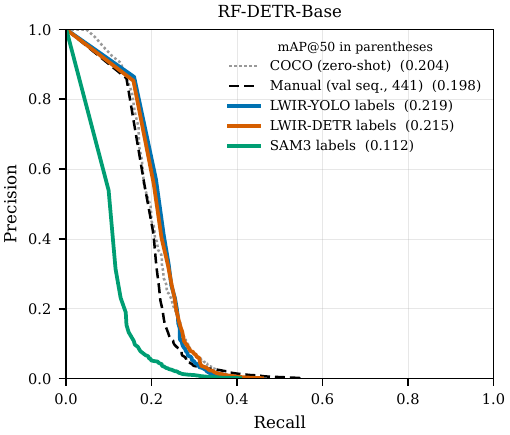}
        \caption{DETR -- PR}
        \label{fig:detr_pr}
    \end{subfigure}
\hspace{0.04\textwidth}
    \begin{subfigure}[b]{0.44\textwidth}
        \centering
        \includegraphics[width=\linewidth,keepaspectratio]{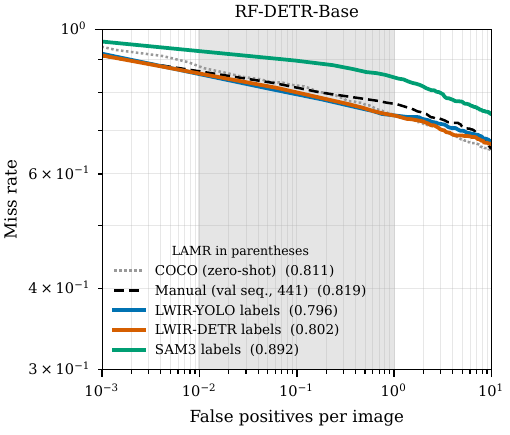}
        \caption{DETR -- MR}
        \label{fig:detr_mr}
    \end{subfigure}

    \caption{Precision--Recall and log-average miss rate (MR) curves for YOLOv11l (top) and RF-DETR-Base (bottom).}
    \label{fig:pr_mr_grid}
\end{figure*}

Using the filtering parameters that yielded the best overall results we construct Table \ref{tab:final_results}, with precision-recall and log-average miss rate curves shown in Figure \ref{fig:pr_mr_grid}. Additionally, qualitative results are shown in Figure \ref{fig:qualitative_grid}, comparing model outputs with the ground truth from our test set. All models were evaluated from stored detections using a single protocol, pycocotools for mAP, max-F1 for P/R/F1, and Caltech LAMR.

\begin{table*}[tbp]
    \centering
    \caption{Low-light RGB detection on the test sequence (1440 images, 1173
    annotated instances). Training-set size is reported per row, varying based on the filtering parameters.
    \textcolor{green}{Green} marks results better than the 441-frame manual
    reference within the same base model, \textcolor{red}{red} worse; color is applied to F1, mAP and LAMR only, since P and R are not informative as standalone values. P, R, F1, and LAMR are computed with an IoU-threshold of 0.5.
    LAMR is log-average miss rate over FPPI $[10^{-2},10^{0}]$, lower is better.}
    \label{tab:final_results_full}
    \setlength{\tabcolsep}{3pt}
    \renewcommand{\arraystretch}{1.0}
    \scriptsize
    \begin{tabular}{lccrrcccccc}
    \toprule
    & & & \multicolumn{2}{c}{\textbf{Training set}} & \multicolumn{6}{c}{\textbf{Test}} \\
    \cmidrule(lr){4-5}\cmidrule(lr){6-11}
    \textbf{Labels} & $c_t$ & $h_t$ & \textbf{Images} & \textbf{Inst.} &
    \textbf{P} & \textbf{R} & \textbf{F1} & \textbf{mAP@50} & \textbf{mAP@50-95} & \textbf{LAMR}$\downarrow$ \\
    \midrule
    \multicolumn{11}{c}{\textbf{Base Model YOLOv11l}} \\
    \midrule
    COCO (zero-shot)  & --   & --   & --    & --     & 0.596 & 0.154 & 0.244 & 0.199 & 0.118 & 0.815 \\
    Manual (val seq.) & --   & --   & 441   & 363    & 0.342 & 0.162 & 0.220 & 0.188 & 0.097 & 0.832 \\
    \addlinespace[2pt]
    LWIR-YOLO & 0.50 & 0.04 & 29513 & 75161  & 0.584 & 0.228 & \textcolor{green}{0.328} & \textcolor{green}{0.231} & \textcolor{red}{0.080} & \textcolor{green}{0.785} \\
    LWIR-YOLO & 0.50 & 0.02 & 47914 & 174370 & 0.625 & 0.217 & \textcolor{green}{0.323} & \textcolor{green}{0.239} & \textcolor{red}{0.085} & \textcolor{green}{0.778} \\
    LWIR-YOLO & 0.25 & 0.10 & 6164  & 10000  & 0.618 & 0.186 & \textcolor{green}{0.286} & \textcolor{green}{0.208} & \textcolor{red}{0.090} & \textcolor{green}{0.802} \\
    LWIR-YOLO & 0.25 & 0.08 & 9489  & 15796  & 0.571 & 0.217 & \textcolor{green}{0.315} & \textcolor{green}{0.226} & \textcolor{red}{0.091} & \textcolor{green}{0.787} \\
    LWIR-YOLO & 0.25 & 0.06 & 16455 & 30716  & 0.688 & 0.222 & \textcolor{green}{0.335} & \textcolor{green}{0.252} & \textcolor{red}{0.091} & \textcolor{green}{0.763} \\
    LWIR-YOLO & 0.25 & 0.04 & 27930 & 65739  & 0.629 & 0.214 & \textcolor{green}{0.319} & \textcolor{green}{0.243} & \textcolor{red}{0.088} & \textcolor{green}{0.777} \\
    LWIR-YOLO & 0.25 & 0.02 & 51483 & 216466 & 0.613 & 0.208 & \textcolor{green}{0.311} & \textcolor{green}{0.282} & \textcolor{green}{0.124} & \textcolor{green}{0.750} \\
    LWIR-YOLO & 0.25 & 0.00 & 54854 & 252418 & 0.556 & 0.218 & \textcolor{green}{0.313} & \textcolor{green}{0.254} & \textcolor{red}{0.088} & \textcolor{green}{0.769} \\
    LWIR-YOLO & 0.10 & 0.04 & 28388 & 65208  & 0.579 & 0.198 & \textcolor{green}{0.295} & \textcolor{green}{0.221} & \textcolor{red}{0.077} & \textcolor{green}{0.795} \\
    LWIR-YOLO & 0.10 & 0.02 & 54990 & 235840 & 0.567 & 0.221 & \textcolor{green}{0.318} & \textcolor{green}{0.258} & \textcolor{red}{0.091} & \textcolor{green}{0.764} \\
    \addlinespace[2pt]
    LWIR-DETR & 0.50 & 0.04 & 27832 & 68102  & 0.503 & 0.217 & \textcolor{green}{0.303} & \textcolor{green}{0.242} & \textcolor{red}{0.082} & \textcolor{green}{0.777} \\
    LWIR-DETR & 0.50 & 0.02 & 50152 & 180477 & 0.621 & 0.232 & \textcolor{green}{0.338} & \textcolor{green}{0.259} & \textcolor{red}{0.092} & \textcolor{green}{0.763} \\
    LWIR-DETR & 0.25 & 0.10 & 5163  & 8698   & 0.646 & 0.149 & \textcolor{green}{0.242} & \textcolor{red}{0.165} & \textcolor{red}{0.068} & \textcolor{red}{0.842} \\
    LWIR-DETR & 0.25 & 0.08 & 8123  & 14082  & 0.698 & 0.166 & \textcolor{green}{0.269} & \textcolor{green}{0.195} & \textcolor{red}{0.076} & \textcolor{green}{0.812} \\
    LWIR-DETR & 0.25 & 0.06 & 13914 & 25340  & 0.646 & 0.195 & \textcolor{green}{0.300} & \textcolor{green}{0.238} & \textcolor{green}{0.101} & \textcolor{green}{0.777} \\
    LWIR-DETR & 0.25 & 0.04 & 24617 & 53862  & 0.560 & 0.217 & \textcolor{green}{0.312} & \textcolor{green}{0.250} & \textcolor{red}{0.087} & \textcolor{green}{0.767} \\
    LWIR-DETR & 0.25 & 0.02 & 53684 & 212098 & 0.522 & 0.213 & \textcolor{green}{0.303} & \textcolor{green}{0.252} & \textcolor{red}{0.089} & \textcolor{green}{0.769} \\
    LWIR-DETR & 0.25 & 0.00 & 65730 & 353696 & 0.599 & 0.228 & \textcolor{green}{0.331} & \textcolor{green}{0.243} & \textcolor{red}{0.086} & \textcolor{green}{0.775} \\
    LWIR-DETR & 0.10 & 0.04 & 20934 & 49841  & 0.544 & 0.204 & \textcolor{green}{0.297} & \textcolor{green}{0.246} & \textcolor{red}{0.088} & \textcolor{green}{0.776} \\
    LWIR-DETR & 0.10 & 0.02 & 58563 & 269275 & 0.486 & 0.210 & \textcolor{green}{0.293} & \textcolor{green}{0.240} & \textcolor{red}{0.078} & \textcolor{green}{0.780} \\
    \addlinespace[2pt]
    SAM3    & 0.50 & 0.04 & 28681 & 70455  & 0.450 & 0.099 & \textcolor{red}{0.162} & \textcolor{red}{0.129} & \textcolor{red}{0.042} & \textcolor{red}{0.877} \\
    SAM3    & 0.50 & 0.02 & 46723 & 137302 & 0.216 & 0.093 & \textcolor{red}{0.130} & \textcolor{red}{0.138} & \textcolor{red}{0.048} & \textcolor{red}{0.867} \\
    SAM3    & 0.25 & 0.04 & 17624 & 39054  & 0.140 & 0.100 & \textcolor{red}{0.117} & \textcolor{red}{0.129} & \textcolor{red}{0.043} & \textcolor{red}{0.879} \\
    SAM3    & 0.25 & 0.02 & 47310 & 195488 & 0.158 & 0.087 & \textcolor{red}{0.112} & \textcolor{red}{0.127} & \textcolor{red}{0.035} & \textcolor{red}{0.877} \\
    SAM3    & 0.10 & 0.04 & 3677  & 7387   & 0.414 & 0.095 & \textcolor{red}{0.155} & \textcolor{red}{0.148} & \textcolor{red}{0.058} & \textcolor{red}{0.860} \\
    SAM3    & 0.10 & 0.02 & 14847 & 80382  & 0.136 & 0.085 & \textcolor{red}{0.105} & \textcolor{red}{0.120} & \textcolor{red}{0.038} & \textcolor{red}{0.885} \\
    \midrule
    \multicolumn{11}{c}{\textbf{Base Model RF-DETR-Base}} \\
    \midrule
    COCO (zero-shot)  & --   & --   & --    & --     & 0.446 & 0.153 & 0.228 & 0.204 & 0.118 & 0.811 \\
    Manual (val seq.) & --   & --   & 441   & 363    & 0.442 & 0.129 & 0.199 & 0.198 & 0.103 & 0.819 \\
    \addlinespace[2pt]
    LWIR-YOLO & 0.50 & 0.04 & 29513 & 75161  & 0.329 & 0.139 & \textcolor{red}{0.195} & \textcolor{green}{0.218} & \textcolor{red}{0.079} & \textcolor{green}{0.797} \\
    LWIR-YOLO & 0.50 & 0.02 & 47914 & 174370 & 0.325 & 0.140 & \textcolor{red}{0.195} & \textcolor{green}{0.228} & \textcolor{red}{0.085} & \textcolor{green}{0.788} \\
    LWIR-YOLO & 0.25 & 0.10 & 6164  & 10000  & 0.260 & 0.055 & \textcolor{red}{0.091} & \textcolor{red}{0.161} & \textcolor{red}{0.071} & \textcolor{red}{0.844} \\
    LWIR-YOLO & 0.25 & 0.08 & 9489  & 15796  & 0.374 & 0.111 & \textcolor{red}{0.171} & \textcolor{red}{0.187} & \textcolor{red}{0.070} & \textcolor{red}{0.821} \\
    LWIR-YOLO & 0.25 & 0.06 & 16455 & 30716  & 0.266 & 0.078 & \textcolor{red}{0.120} & \textcolor{green}{0.210} & \textcolor{red}{0.077} & \textcolor{green}{0.803} \\
    LWIR-YOLO & 0.25 & 0.04 & 27930 & 65739  & 0.279 & 0.115 & \textcolor{red}{0.163} & \textcolor{green}{0.225} & \textcolor{red}{0.084} & \textcolor{green}{0.792} \\
    LWIR-YOLO & 0.25 & 0.02 & 51483 & 216466 & 0.282 & 0.103 & \textcolor{red}{0.151} & \textcolor{green}{0.219} & \textcolor{red}{0.080} & \textcolor{green}{0.796} \\
    LWIR-YOLO & 0.25 & 0.00 & 54854 & 252418 & 0.283 & 0.085 & \textcolor{red}{0.131} & \textcolor{green}{0.224} & \textcolor{red}{0.083} & \textcolor{green}{0.794} \\
    LWIR-YOLO & 0.10 & 0.04 & 28388 & 65208  & 0.313 & 0.117 & \textcolor{red}{0.170} & \textcolor{green}{0.216} & \textcolor{red}{0.085} & \textcolor{green}{0.801} \\
    LWIR-YOLO & 0.10 & 0.02 & 54990 & 235840 & 0.295 & 0.112 & \textcolor{red}{0.162} & \textcolor{green}{0.226} & \textcolor{red}{0.089} & \textcolor{green}{0.791} \\
    \addlinespace[2pt]
    LWIR-DETR & 0.50 & 0.04 & 27832 & 68102  & 0.462 & 0.152 & \textcolor{green}{0.228} & \textcolor{green}{0.211} & \textcolor{red}{0.084} & \textcolor{green}{0.801} \\
    LWIR-DETR & 0.50 & 0.02 & 50152 & 180477 & 0.565 & 0.206 & \textcolor{green}{0.302} & \textcolor{green}{0.228} & \textcolor{red}{0.091} & \textcolor{green}{0.789} \\
    LWIR-DETR & 0.25 & 0.10 & 5163  & 8698   & 0.422 & 0.103 & \textcolor{red}{0.165} & \textcolor{red}{0.161} & \textcolor{red}{0.067} & \textcolor{red}{0.844} \\
    LWIR-DETR & 0.25 & 0.08 & 8123  & 14082  & 0.156 & 0.052 & \textcolor{red}{0.078} & \textcolor{red}{0.190} & \textcolor{red}{0.074} & \textcolor{green}{0.818} \\
    LWIR-DETR & 0.25 & 0.06 & 13914 & 25340  & 0.412 & 0.124 & \textcolor{red}{0.190} & \textcolor{green}{0.207} & \textcolor{red}{0.076} & \textcolor{green}{0.805} \\
    LWIR-DETR & 0.25 & 0.04 & 24617 & 53862  & 0.461 & 0.126 & \textcolor{red}{0.198} & \textcolor{green}{0.228} & \textcolor{red}{0.080} & \textcolor{green}{0.788} \\
    LWIR-DETR & 0.25 & 0.02 & 53684 & 212098 & 0.437 & 0.140 & \textcolor{green}{0.213} & \textcolor{green}{0.215} & \textcolor{red}{0.085} & \textcolor{green}{0.802} \\
    LWIR-DETR & 0.25 & 0.00 & 65730 & 353696 & 0.376 & 0.116 & \textcolor{red}{0.177} & \textcolor{green}{0.221} & \textcolor{red}{0.078} & \textcolor{green}{0.797} \\
    LWIR-DETR & 0.10 & 0.04 & 20934 & 49841  & 0.371 & 0.100 & \textcolor{red}{0.157} & \textcolor{green}{0.227} & \textcolor{red}{0.082} & \textcolor{green}{0.790} \\
    LWIR-DETR & 0.10 & 0.02 & 58563 & 269275 & 0.276 & 0.104 & \textcolor{red}{0.151} & \textcolor{green}{0.216} & \textcolor{red}{0.079} & \textcolor{green}{0.799} \\
    \addlinespace[2pt]
    SAM3    & 0.50 & 0.04 & 28681 & 70455  & 0.379 & 0.090 & \textcolor{red}{0.145} & \textcolor{red}{0.118} & \textcolor{red}{0.039} & \textcolor{red}{0.885} \\
    SAM3    & 0.50 & 0.02 & 46723 & 137302 & 0.320 & 0.051 & \textcolor{red}{0.088} & \textcolor{red}{0.112} & \textcolor{red}{0.044} & \textcolor{red}{0.891} \\
    SAM3    & 0.25 & 0.04 & 17624 & 39054  & 0.257 & 0.084 & \textcolor{red}{0.126} & \textcolor{red}{0.131} & \textcolor{red}{0.041} & \textcolor{red}{0.876} \\
    SAM3    & 0.25 & 0.02 & 47310 & 195488 & 0.175 & 0.079 & \textcolor{red}{0.109} & \textcolor{red}{0.112} & \textcolor{red}{0.040} & \textcolor{red}{0.892} \\
    SAM3    & 0.10 & 0.04 & 3677  & 7387   & 0.297 & 0.113 & \textcolor{red}{0.164} & \textcolor{red}{0.163} & \textcolor{red}{0.080} & \textcolor{red}{0.847} \\
    SAM3    & 0.10 & 0.02 & 14847 & 80382  & 0.280 & 0.106 & \textcolor{red}{0.154} & \textcolor{red}{0.160} & \textcolor{red}{0.052} & \textcolor{red}{0.849} \\
    \midrule
    \multicolumn{11}{c}{\textbf{Base Model SAM3}} \\
    \midrule
    Zero-shot (RGB)   & --   & --   & --    & --     & 0.381 & 0.125 & 0.189 & 0.127 & 0.046 & 0.877 \\
    \bottomrule
    \end{tabular}
\end{table*}

\begin{table*}
    \centering
    \caption{Low-light RGB detection on the test sequence (1440 images, 1173
    annotated instances). All autolabeled models use the same filtration setting, $c_t = 0.25$ and $h_t = 0.02$. Resulting training set size is reported per row.
    \textcolor{green}{Green} marks results better than the 441-frame manual
    reference with the same base model, \textcolor{red}{red} worse; color is applied to F1, mAP and LAMR only, since P and R are not informative as standalone values. P, R, F1, and LAMR are computed with an IoU-threshold of 0.5.
    LAMR is log-average miss rate over FPPI $[10^{-2},10^{0}]$, lower is better.}
    \label{tab:final_results}
    \setlength{\tabcolsep}{4pt}
    \renewcommand{\arraystretch}{1.1}
    \footnotesize
    \begin{tabular}{lccrrcccccc}
    \toprule
    & & & \multicolumn{2}{c}{\textbf{Training set}} & \multicolumn{6}{c}{\textbf{Test}} \\
    \cmidrule(lr){4-5}\cmidrule(lr){6-11}
    \textbf{Labels} & $c_t$ & $h_t$ & \textbf{Images} & \textbf{Inst.} &
    \textbf{P} & \textbf{R} & \textbf{F1} & \textbf{mAP@50} & \textbf{mAP@50-95} & \textbf{LAMR}$\downarrow$ \\
    \midrule
    \multicolumn{11}{c}{\textbf{YOLOv11l}} \\
    \midrule
    COCO (zero-shot)  & --   & --   & --    & --     & 0.596 & 0.154 & 0.244 & 0.199 & 0.118 & 0.815 \\
    Manual (val seq.) & --   & --   & 441   & 363    & 0.342 & 0.162 & 0.220 & 0.188 & 0.097 & 0.832 \\
    \addlinespace[2pt]
    LWIR-YOLO         & 0.25 & 0.02 & 51483 & 216466 & 0.613 & 0.208 & \textcolor{green}{0.311} & \textcolor{green}{0.282} & \textcolor{green}{0.124} & \textcolor{green}{0.750} \\
    LWIR-DETR         & 0.25 & 0.02 & 53684 & 212098 & 0.522 & 0.213 & \textcolor{green}{0.303} & \textcolor{green}{0.252} & \textcolor{red}{0.089} & \textcolor{green}{0.769} \\
    SAM3              & 0.25 & 0.02 & 47310 & 195488 & 0.158 & 0.087 & \textcolor{red}{0.112} & \textcolor{red}{0.127} & \textcolor{red}{0.035} & \textcolor{red}{0.877} \\
    \midrule
    \multicolumn{11}{c}{\textbf{RF-DETR-Base}} \\
    \midrule
    COCO (zero-shot)  & --   & --   & --    & --     & 0.446 & 0.153 & 0.228 & 0.204 & 0.118 & 0.811 \\
    Manual (val seq.) & --   & --   & 441   & 363    & 0.442 & 0.129 & 0.199 & 0.198 & 0.103 & 0.819 \\
    \addlinespace[2pt]
    LWIR-YOLO         & 0.25 & 0.02 & 51483 & 216466 & 0.282 & 0.103 & \textcolor{red}{0.151} & \textcolor{green}{0.219} & \textcolor{red}{0.080} & \textcolor{green}{0.796} \\
    LWIR-DETR         & 0.25 & 0.02 & 53684 & 212098 & 0.437 & 0.140 & \textcolor{green}{0.213} & \textcolor{green}{0.215} & \textcolor{red}{0.085} & \textcolor{green}{0.802} \\
    SAM3              & 0.25 & 0.02 & 47310 & 195488 & 0.175 & 0.079 & \textcolor{red}{0.109} & \textcolor{red}{0.112} & \textcolor{red}{0.040} & \textcolor{red}{0.892} \\
    \midrule
    \multicolumn{11}{c}{\textbf{SAM3}} \\
    \midrule
    Zero-shot (RGB)   & --   & --   & --    & --     & 0.381 & 0.125 & 0.189 & 0.127 & 0.046 & 0.877 \\
    \bottomrule
    \end{tabular}
\end{table*}

\begin{figure*}[tbp]
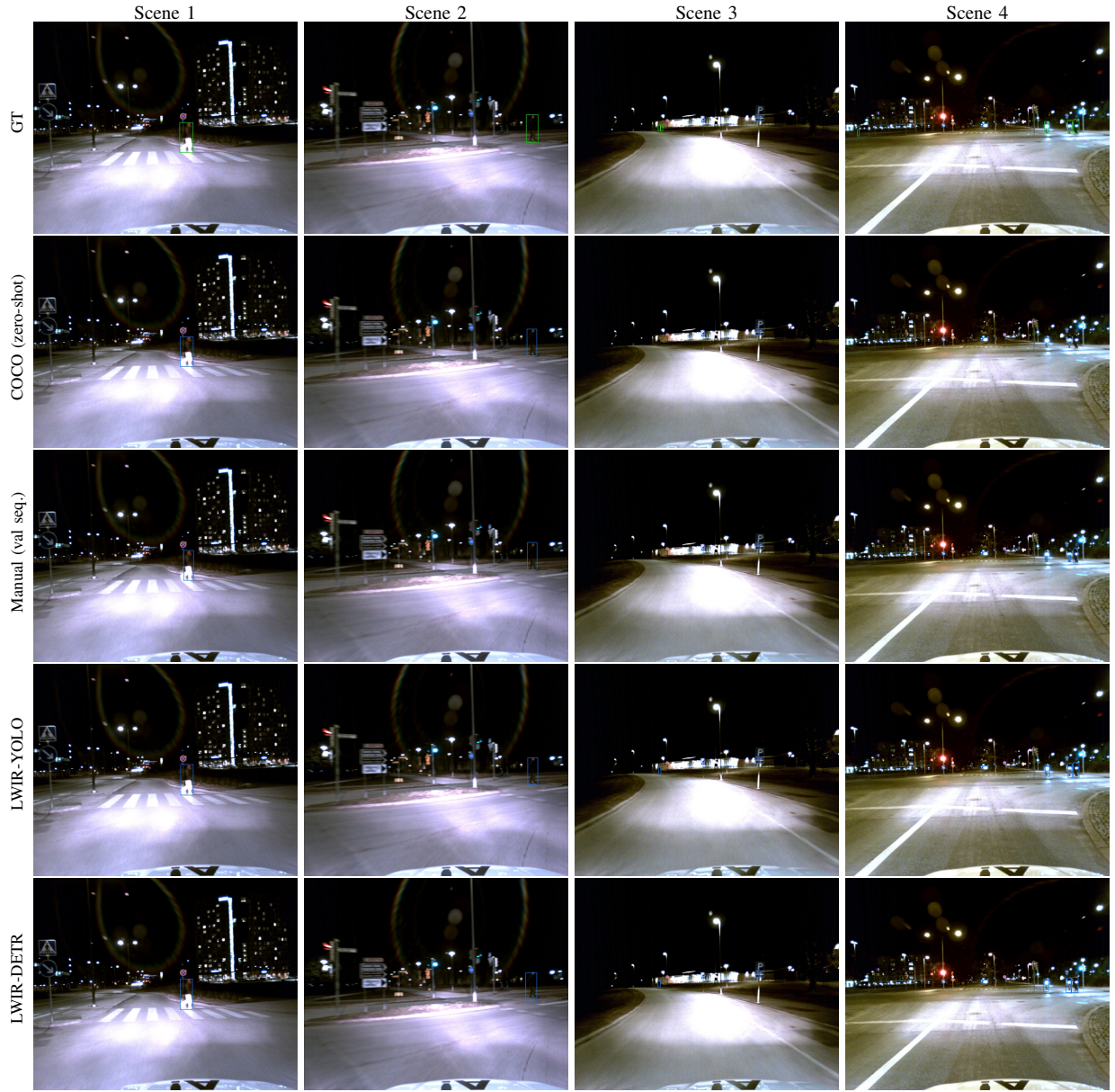

    \centering
    \setlength{\tabcolsep}{1.5pt}
    \renewcommand{\arraystretch}{0.4}
    \newlength{\cellh}\setlength{\cellh}{0.121\textheight}
    \newcommand{\gridcell}[2]{\includegraphics[height=\cellh,keepaspectratio]{#1\detokenize{#2}}}
    \newcommand{\rowlab}[1]{\raisebox{0.5\cellh}[0pt][0pt]{\rotatebox[origin=c]{90}{\scriptsize #1}}}
    \footnotesize
    \begin{tabular}{@{}l cccc @{}}
        & Scene 1 & Scene 2 & Scene 3 & Scene 4 \\
        \rowlab{GT}
            & \gridcell{\sceneA}{00_GT.jpg} & \gridcell{\sceneB}{00_GT.jpg} & \gridcell{\sceneC}{00_GT.jpg} & \gridcell{\sceneD}{00_GT.jpg} \\
        \rowlab{COCO (zero-shot)}
            & \gridcell{\sceneA}{01_yolo_zeroshot.jpg} & \gridcell{\sceneB}{01_yolo_zeroshot.jpg} & \gridcell{\sceneC}{01_yolo_zeroshot.jpg} & \gridcell{\sceneD}{01_yolo_zeroshot.jpg} \\
        \rowlab{Manual (val seq.)}
            & \gridcell{\sceneA}{02_yolo_manual441.jpg} & \gridcell{\sceneB}{02_yolo_manual441.jpg} & \gridcell{\sceneC}{02_yolo_manual441.jpg} & \gridcell{\sceneD}{02_yolo_manual441.jpg} \\
        \rowlab{LWIR-YOLO}
            & \gridcell{\sceneA}{03_yolo_lwiryolo.jpg} & \gridcell{\sceneB}{03_yolo_lwiryolo.jpg} & \gridcell{\sceneC}{03_yolo_lwiryolo.jpg} & \gridcell{\sceneD}{03_yolo_lwiryolo.jpg} \\
        \rowlab{LWIR-DETR}
            & \gridcell{\sceneA}{04_yolo_lwirdetr.jpg} & \gridcell{\sceneB}{04_yolo_lwirdetr.jpg} & \gridcell{\sceneC}{04_yolo_lwirdetr.jpg} & \gridcell{\sceneD}{04_yolo_lwirdetr.jpg} \\
    \end{tabular}
    \caption{Qualitative detection results on test set for YOLOv11l across four scenes (columns) for ground truth and selected models (rows). The LWIR-YOLO and LWIR-DETR labels were generated with $c_t = 0.25$ and $h_t = 0.02$.}
    \label{fig:qualitative_grid}
\end{figure*}





\section{Discussion}

The main contribution of this work is a fully automated pipeline for low-light pedestrian detection tasks that does not require a complicated hardware setup. The value this pipeline brings is that it enables large scale labeling with no human effort. Our results demonstrate that the efficiency translates into performance gains. Models trained on our generated labels beat both off-the-shelf models and the same models fine tuned on a hand labeled subset, on F1, mAP50, and LAMR. While the labels of the autolabeling pipeline do not match human labeling in accuracy, they have the benefit of scaling to large-scale datasets with minimal effort. We show that detector performance can be increased with our autolabels, outperforming a model trained on a reasonable amount of human-labeled data. Additionally, our results show that the performance is not tied to a single labeling model as both LWIR-DETR and LWIR-YOLO producing models that beat the baselines showing that the method works with LWIR dedicated detectors in general. 

The most direct implication of our work is dataset generation. A primary limitation in the process of creating large scale datasets is labeling overhead. Additionally, publicly available datasets are geographically limited to the region of the world where they were collected and can be a mismatch when it comes to training models due to the domain gap as a result of the sensor selection.
The stereo based setup lowers hardware complexity which is the main limitation in an autolabeling system. This makes the proposed pipeline practically applicable enabling a wide range of users to adopt the method. 

Our results also point towards the value of dataset scale. As the RGB pedestrian detection task in low-light conditions is difficult especially in cases such as our dataset where pedestrians are often located at a larger distance from the vehicle providing even less visual information, a small labeled subset cannot provide enough training samples to substantially improve the model. More specifically in our experiments we noticed that baseline models fine tuned on our validation data dropped in performance when compared to their own baseline. This drop in performance showcases that in difficult applications small datasets cannot provide enough data samples to reliably improve model performance. 

An important discussion topic is the limitation posed by a stereo setup. The proposed pipeline is directly affected by two parameters, one being LWIR model performance and second being label transfer accuracy. Stereo setups are efficient to setup compared to other more complicated methods such as the beam splitter used in KAIST. Yet they come with drawbacks, three primary ones to be precise. First being the fact that stereo setups cannot completely cover up the physical gap between the sensors through software. Second being the distance at which pedestrians appear, as the stereo system uses a fixed convergence distance and pedestrians will in most cases not match it. Lastly, the timing limitation, in our setup the LWIR camera did not provide "fire on command" capabilities thus limiting the temporal alignment, better timing approaches that would narrow this gap have the potential to further improve this method. The effects of poor label transfer can be seen on mAP@50-95 as all the other metrics improve compared to baseline it remains low with one exception only, in our best overall model. 
Although the trade-off between an increase in mAP@50 and a decrease in mAP50-95 is still beneficial as the ability to detect a VRU is more important than detecting them with pixel perfect accuracy, this trade-off remains an open improvement point for future research. 

Consequently, an improvement direction for this autolabeling pipeline would involve an improved calibration method tailored for the task. A method that could benefit the pipeline could involve per-detection online calibration, where the modality invariant features of pedestrians in LWIR would be matched to those in RGB leveraging the assumption that the bounding box provides a good location estimation.

The other important limitation comes from the autolabelers themselves. Our generated labels are limited by the autolabelers ability to detect pedestrians in LWIR images. This can be seen as the autolabeler performance on the LWIR validation set carried on to the RGB testing as the labels generated by the best labeler (LWIR-YOLO) resulted in the best performing RGB models. To improve this crucial part of the pipeline better detectors are required. One potential future path to be explored would be LWIR-RGB autolabeling by combining the two available modalities. In addition to this the approach taken in the paper can be also expanded by training and improving LWIR detection models directly.

\section{Conclusion}

In this paper we presented an automated pipeline for low-light pedestrian labeling that does not rely on a complex hardware setup. We replaced the requirement of a beam splitter our previous experiments had with a stereo camera setup, and evaluated the full method end to end from stereo calibration to data collection and the labeling pipeline itself. The evaluation was performed on a dataset collection in the capital region of Finland. 

Our results show that the generated labels are useful and result in improved models. The models fine tuned on our generated labels beat both off-the-shelf models and those same models fine tuned on hand labeled subsets of our data. The results do not indicate that automatic labels match human labels on a per image basis but that autolabeling at scale yields better results than reasonable amounts human effort. The results were consisted with both dedicated LWIR autolabelers tested which indicates that the method is not bounded by a specific model selection. These findings are further proof that automatic low-light pedestrian labeling can be accessible from a hardware standpoint and effective in performance. 

The approach has potential for improvement. A primary limitation is the label transfer accuracy which affects localization and is visible in the mAP@50-95 performance drop. A promising direction appears to be label transfer system tailor to the task, such as a per-detection based label transfer system matching pedestrian features across modalities. A second direction is the improvement of the labeling models, approaches can vary by leveraging both available modalities and diverging to ward RGB and LWIR labeling or directly improving LWIR detection capabilities which will have the downstream effect of improving RGB model performance.

\bibliographystyle{ieeetr}
\bibliography{bibliography}

\end{document}